%% file: main.tex
\pdfoutput=1
\documentclass{article}
\usepackage{iclr2027_conference,times}
\usepackage[T1]{fontenc}

\input{math_commands.tex}

\PassOptionsToPackage{hyphens}{url}
\usepackage{hyperref}
\usepackage{url}
\usepackage{amsmath}
\usepackage{amssymb}
\usepackage{booktabs}
\usepackage{graphicx}
\usepackage{multicol}
\usepackage{xcolor}
\usepackage{xspace}
\usepackage{listings}
\usepackage{tabularx}
\usepackage{colortbl}
\usepackage{makecell}
\usepackage{pifont}
\newcommand{\cmark}{\ding{51}}
\newcommand{\xmark}{\ding{55}}
\definecolor{oursrow}{HTML}{EAF0F8}
\usepackage{placeins}
\usepackage{tikz}
\usetikzlibrary{arrows.meta}
\definecolor{figink}{HTML}{1A1A19}
\definecolor{figink2}{HTML}{52514E}
\definecolor{figpanel}{HTML}{F4F4F1}
\definecolor{figrule}{HTML}{C9C8C2}
\definecolor{figcell}{HTML}{DAD9D3}
\definecolor{figtint}{HTML}{E6ECF6}
\definecolor{figblue}{HTML}{3B6FB6}
\definecolor{figorange}{HTML}{D46E2C}
\definecolor{figaqua}{HTML}{2A9D8F}
\definecolor{figpurple}{HTML}{7B5EA7}
\definecolor{figgrey}{HTML}{A19F98}
\definecolor{figgreytag}{HTML}{85837C}
\definecolor{figlatent}{HTML}{2E2E2B}
\input{taskcard}

\defcitealias{dcwp2024data}{DCWP, 2024a}
\defcitealias{dcwp2024minpay}{DCWP, 2024b}

\newcommand{\ours}{Argo-Bench\xspace}
\input{sec/numbers}

\lstdefinestyle{prompt}{
  basicstyle=\ttfamily\footnotesize,
  breaklines=true,
  columns=fullflexible,
  keepspaces=true,
  frame=single,
  rulecolor=\color{black!40},
  framesep=6pt,
  xleftmargin=6pt,
  xrightmargin=6pt,
  escapeinside={(*@}{@*)},
  upquote=true,
}

\definecolor{toolhead}{gray}{0.93}
\definecolor{toollabel}{gray}{0.45}
\newcommand{\toolcard}[5]{%
  \par\noindent
  \begin{minipage}{\linewidth}
  \renewcommand{\arraystretch}{1.25}\frenchspacing%
  \begin{tabularx}{\linewidth}{@{\hspace{6pt}}>{\raggedright\arraybackslash\sffamily\scriptsize\leavevmode\color{toollabel}}p{1.9cm}>{\raggedright\arraybackslash\small}X@{\hspace{6pt}}}
    \arrayrulecolor{black!40}\hline
    \rowcolor{toolhead}\multicolumn{2}{@{\hspace{6pt}}p{\dimexpr\linewidth-12pt\relax}@{\hspace{6pt}}}{%
      \rule{0pt}{2.4ex}\textbf{\texttt{#1}}\hfill{\sffamily\scriptsize\color{toollabel}\MakeUppercase{#2} server}}\\
    \hline
    SIGNATURE   & \texttt{#3}\\
    DESCRIPTION & #4\\
    EXAMPLE     & \texttt{#5}\\[2pt]
    \hline
  \end{tabularx}
  \end{minipage}\par\vspace{10pt}}

\title{Argo-Bench: Evaluating Data Agents on Enterprise-Scale Workflows}

\author{Gabriel Tomitsuka, Arman Raayatsanati, Emma Xing, Duke Gand, Joseph J Ma\\
TextQL\\
\texttt{\{gabriel,arman,emma,duke,joseph\}@textql.com}}

\iclrfinalcopy
\begin{document}

\maketitle
\lhead{Preprint. Under review.}

\input{sec/00_abstract}
\input{sec/01_introduction}
\input{sec/02_benchmark}
\input{sec/03_evaluation}
\input{sec/04_related_work}
\input{sec/05_limitations}
\input{sec/06_conclusion}
\input{sec/07_statements}
\input{sec/08_acknowledgments}

\input{main.bbl}
\clearpage
\appendix
\input{sec/appendix/A_donor_datasets}
\input{sec/appendix/B_minimum_pay}
\input{sec/appendix/C_warehouse_tables}

\input{sec/appendix/D_reward_hacking}
\input{sec/appendix/E_system_prompt}
\input{sec/appendix/F_agent_tools}
\input{sec/appendix/G_grading}
\input{sec/appendix/H_spider2}
\input{sec/appendix/I_findings}
\input{sec/appendix/J_task_audit}
\input{sec/appendix/K_warehouse_cost}
\input{sec/appendix/L_example_tasks}

\end{document}

%% file: math_commands.tex
\usepackage{amsmath,amsfonts,bm}

\def\eqref#1{equation~\ref{#1}}

\def\1{\bm{1}}

\DeclareMathAlphabet{\mathsfit}{\encodingdefault}{\sfdefault}{m}{sl}
\SetMathAlphabet{\mathsfit}{bold}{\encodingdefault}{\sfdefault}{bx}{n}

%% file: taskcard.tex
\definecolor{cdkw}{HTML}{1F4E9C}
\definecolor{cdfn}{HTML}{5A4A9C}
\definecolor{cdstr}{HTML}{A31515}
\definecolor{cdcmt}{HTML}{6E7781}
\definecolor{cdtbl}{HTML}{0B6E4F}
\definecolor{cdbg}{HTML}{F7F7F4}
\definecolor{cardrule}{gray}{0.55}
\definecolor{cardhead}{gray}{0.93}
\definecolor{keyfill}{HTML}{FFF3D6}
\definecolor{chipgood}{HTML}{DCEFDF}
\definecolor{chippart}{HTML}{FBEBC8}
\definecolor{chipzero}{gray}{0.92}
\definecolor{chipref}{HTML}{0B6E4F}

\ifcsname\string\color@dmbg\endcsname
  \definecolor{cdkw}{HTML}{8AB4F8}
  \definecolor{cdfn}{HTML}{C3B1F2}
  \definecolor{cdstr}{HTML}{F0A58C}
  \definecolor{cdcmt}{HTML}{9AA0A6}
  \definecolor{cdtbl}{HTML}{7BD3A5}
  \definecolor{cdbg}{HTML}{27272B}
  \definecolor{cardrule}{gray}{0.6}
  \definecolor{cardhead}{HTML}{333338}
  \definecolor{keyfill}{HTML}{4A3F22}
  \definecolor{chipgood}{HTML}{234A2C}
  \definecolor{chippart}{HTML}{5A4A1A}
  \definecolor{chipzero}{HTML}{3A3A3E}
\fi

\lstdefinelanguage{bqsql}{
  morekeywords={SELECT,FROM,WHERE,JOIN,LEFT,ON,USING,GROUP,BY,ORDER,AS,WITH,AND,OR,
    IN,IS,NOT,NULL,BETWEEN,DESC,ASC,DISTINCT},
  morekeywords=[2]{COUNT,COUNTIF,SUM,DATE,DATETIME_DIFF,EXTRACT,DAYOFWEEK,CAST,
    SAFE_CAST,INT64,NUMERIC,SECOND,IFNULL,COALESCE,FORMAT_DATETIME,ROUND,
    APPROX_QUANTILES,OFFSET,ST_DISTANCE,ST_GEOGPOINT},
  sensitive=true,
  morecomment=[l]{--},
  morestring=[b]',
  morestring=[b]",
}
\lstdefinestyle{card}{
  basicstyle=\fontsize{6.3}{6.7}\selectfont\ttfamily,
  columns=fixed,
  basewidth=0.53em,
  keepspaces=true,
  showstringspaces=false,
  upquote=true,
  alsoletter={_},
  keywordstyle=\color{cdkw}\bfseries,
  keywordstyle=[2]\color{cdfn},
  stringstyle=\color{cdstr},
  commentstyle=\color{cdcmt}\itshape,
  emphstyle=\color{cdtbl}\bfseries,
  numbers=left,
  numberstyle=\fontsize{5}{6}\selectfont\ttfamily\color{cdcmt},
  numbersep=6pt,
  xleftmargin=2.0em,
  framexleftmargin=2.0em,
  framesep=3pt,
  frame=single,
  framerule=0pt,
  rulecolor=\color{cdbg},
  backgroundcolor=\color{cdbg},
  aboveskip=3pt,
  belowskip=2pt,
  breaklines=false,
}
\lstdefinestyle{cardsql}{
  style=card, language=bqsql,
  emph={XX_PAYOUT_PERIODS,OE_ORDER_HEADERS_ALL,XX_DRIVER_SHIFTS,AP_INVOICES_ALL,
    AP_INVOICE_LINES_ALL,XX_PAY_PERIOD_INCENTIVES,GL_IMPORT_REFERENCES,XLA_AE_LINES,
    XX_GL_INTERFACE_HIST,XX_REFUNDS,XX_DELIVERY_LEGS},
}
\lstdefinestyle{cardpy}{
  style=card, language=Python,
  morekeywords={as},
  emph={MissionControl,file_forecasts,publish_data_sources,ban_couriers,ban_customers,summary,note},
}
\newcommand{\cardsql}[1]{\lstinputlisting[style=cardsql]{#1}}
\newcommand{\cardpy}[1]{\lstinputlisting[style=cardpy]{#1}}

\newcommand{\code}[1]{\texttt{#1}}

\newcommand{\cardbegin}[3]{%
  \par\noindent
  \colorbox{cardhead}{\parbox{\dimexpr\linewidth-2\fboxsep\relax}{%
    \rule{0pt}{2.6ex}\textbf{\texttt{#1}}\hfill
    {\sffamily\scriptsize\color{toollabel}\MakeUppercase{#2}\enspace$\cdot$\enspace #3}}}%
  \par}

\newcommand{\cardsection}[1]{%
  \par\vspace{5pt}\noindent{\color{cardrule}\rule{\linewidth}{0.4pt}}\par
  \vspace{1pt}\noindent\textbf{\textit{#1}}\par\nopagebreak\vspace{2.5pt}}

\newcommand{\cardlabel}[2]{%
  \par\vspace{3pt}\noindent
  {\sffamily\scriptsize\color{toollabel}#1}\enspace{\footnotesize #2}\par\nopagebreak}

\newcommand{\cardstep}[3]{%
  \par\vspace{2pt}\noindent
  {\sffamily\scriptsize\color{toollabel}STEP #1}\enspace
  {\small\textbf{\texttt{#2}}}\enspace{\footnotesize #3}\par\nopagebreak}

\newsavebox{\cardpromptbox}
\newenvironment{cardprompt}{%
  \par\noindent\begin{lrbox}{\cardpromptbox}%
  \begin{minipage}[t]{\dimexpr\linewidth-10pt\relax}\footnotesize\setlength{\parskip}{3pt}}{%
  \end{minipage}\end{lrbox}%
  \noindent\hbox{{\color{black!25}\vrule width 2pt}\hskip 8pt\usebox{\cardpromptbox}}\par}

\newcommand{\chipref}[2]{%
  \colorbox{chipref}{\color{white}\scriptsize\sffamily #1\ \textbf{#2}}\hspace{3pt plus 3pt}}
\newcommand{\chip}[3]{%
  \colorbox{#1}{\scriptsize\sffamily #2\ \textbf{#3}}\hspace{3pt plus 3pt}}
\newcommand{\chiprow}[1]{\par\vspace{5pt}\noindent{\raggedright\setlength{\fboxsep}{2.5pt}%
  \setlength{\baselineskip}{13pt}#1\par}\vspace{1pt}}

\makeatletter
\newcommand{\cardcaption}[2]{%
  \refstepcounter{table}\label{#1}%
  \par\vspace{2pt}{\@makecaption{\tablename~\thetable}{#2}}\par\vspace{4pt}}
\makeatother

%% file: sec/numbers.tex
\newcommand{\ntasks}{210\xspace}
\newcommand{\npartial}{99\xspace}

\newcommand{\npromptwords}{159\xspace}
\newcommand{\nacted}{178\xspace}

%% file: sec/00_abstract.tex
\begin{abstract}
Real-world enterprise data science and analytics workflows require reasoning across dozens of tables, performing statistical analyses, and acting on the results. Established text-to-SQL benchmarks evaluate query generation alone, and audits have found their answer keys frequently wrong. Because real enterprise warehouses are too sensitive to release, these benchmarks are built on public datasets where a business event fits in a single table. We introduce \ours, an evaluation framework comprising \ntasks data science and analytics tasks. Drawing on public data, peer-reviewed industry literature, and regulatory filings, we simulate a food delivery platform in New York City at true scale, with 81 million orders in 2024, grounded economics, fraud patterns, and marketplace incentives. We export this world to an ERP warehouse of 235 tables and 7.5 billion rows, modeled on the Oracle E-Business Suite schema. The simulator's ground-truth state is withheld from the warehouse the agent sees, so tasks require reconstructing facts by navigating the warehouse before acting on them. \ours goes beyond text-to-SQL: the agent files actions such as banning fraudulent accounts, allocating courier incentive budgets, or issuing back pay, and the grader scores each by its consequences in the simulator. Every task has an executable reference solution that demonstrates solvability using only the warehouse. The strongest of 14 frontier and open-weight models scores 95 or higher on only 34.8\% of tasks and averages 59.5 points. We hope \ours drives progress toward agents that understand, navigate, and act within real data environments.
\end{abstract}

%% file: sec/01_introduction.tex
\section{Introduction}
\label{sec:intro}
\suppressfloats[t]

\begin{figure}[t]
  \centering
  \includegraphics{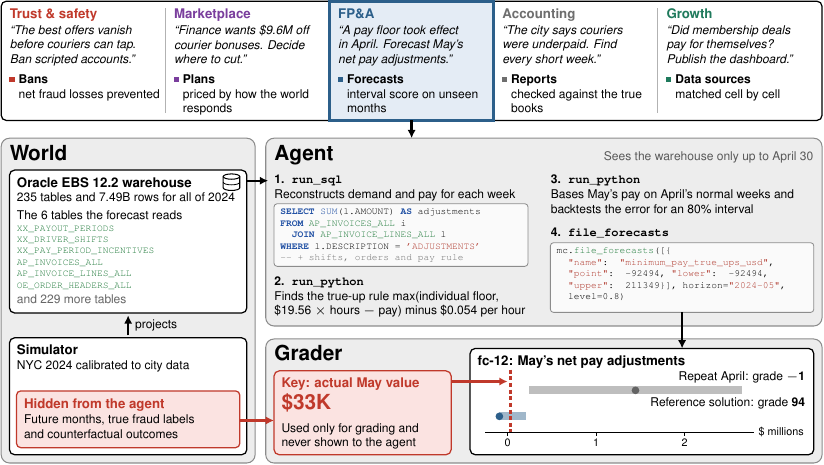}
  \caption{\ours grades the filings of an agent with an incomplete view of the world. Top: one task from each of the five business areas (as in Figure~\ref{fig:tasks}). Bottom: the forecasting task fc-12 (Appendix~\ref{app:examples}). The agent sees the warehouse only up to April 30, and only the grader sees May's actual value. The reference solution reads 6 of the 235 tables, recovers the true-up rule, files a forecast for May with an 80\% interval, and grades 94 out of 100, while repeating April's \$1.44 million misses by \$1.41 million and grades $-1$.}
  \label{fig:overview}
\end{figure}

In recent years, data agents have advanced from answering simple text-to-SQL questions over small, well-documented schemas to performing long-horizon data science tasks, publishing durable company-wide assets such as dashboards, and acting on their findings, for instance, by adjusting promotions and banning fraudulent accounts~\citep{sun2025data,liu2025supporting,li2026deepeye}. Rising scores on established data benchmarks could be read as a sign that enterprise data work is close to solved. The best Spider 2.0-Snow score has risen from 23.8\% at release~\citep{lei2025spider} to 96.7\%, and the top BIRD entry reaches 82.4\% against a human 93.0\%.\footnote{Spider 2.0 leaderboard: \url{https://spider2-sql.github.io/}; BIRD leaderboard: \url{https://bird-bench.github.io/}; both accessed September 24, 2026.}

However, current benchmarks primarily evaluate text-to-SQL performance and are not representative of enterprise agentic data workflows in three important ways. First, they run on a patchwork of public data. Spider 2.0-Snow's 547 questions sparsely cover a sprawling collection of 152 databases (Appendix~\ref{app:spider2}). Over a third of those databases are used for a single question only. Two thirds of the questions use BigQuery public datasets; another quarter use local sample databases~\citep{lei2025spider}. These public and local sources' schemas are documented in tutorials and textbooks. Furthermore, there is a high degree of overlap among tables: at least two thirds of the tables are date-, geography-, or version-sharded copies of another table. In an enterprise warehouse, different modules must agree on the same numbers, and one business event may touch ten or more tables~\citep{plattner2014impact}.

Second, they are not end-to-end. Many of the most valuable enterprise data science tasks cannot be done in SQL. Fitting a demand forecast, training a fraud model, or optimizing a courier schedule requires statistical, machine learning, and optimization libraries. Additionally, findings lead to decisions, which are ultimately judged by their return on investment. For instance, a promotion with a high redemption rate can still lose money if most of those orders would have been placed anyway~\citep{xu2025causal}. A gold answer cannot distinguish between such decisions.

Third, they are not verifiable. On real data, a benchmark can only measure agreement with its annotators since writing the answer key requires solving the task. Even when the answer is in the data, annotators may miss it. In a recent audit, 62.8\% of Spider 2.0-Snow's released gold queries and 52.8\% of BIRD Mini-Dev's were found to be erroneous, most often because annotators misread the data or the schema, and correcting BIRD's errors moved agents' leaderboard standings by up to nine places~\citep{jin2026pervasive}. Furthermore, for many of the most valuable enterprise tasks, the answer may not be in the data at all. Fraud left undetected leaves no label~\citep{altman2023realistic}, and the outcomes of a rejected decision are never observed.

Large enterprises keep the records that financial planning, forecasting, and fraud detection require in enterprise resource planning (ERP) systems~\citep{davenport1998putting} such as Oracle E-Business Suite (EBS), SAP S/4HANA, and Oracle Fusion Cloud, extended with custom tables representing idiosyncrasies and workflows unique to the business~\citep{brehm2001tailoring}. Because the same systems hold the company's ledgers, payroll, and customer records, access to them is heavily restricted. For this reason, ERP data has remained practically unexplored in benchmarks to date. When such data is released, it must be anonymized, which can break the relational structure that answers depend on. In an early release of BEAVER~\citep{chen2024beaver}, the closest attempt to date, \citet{chung2025long} report primary keys that violate their uniqueness constraints and questions whose gold query is null.

We instead simulate a business and grade against the simulation's ground truth (Figure~\ref{fig:overview}). \ours models a food delivery platform in New York City in 2024, a three-sided marketplace whose economics are disclosed to the city every month~\citepalias{dcwp2024data}. The simulator reflects a real minimum-pay increase in April 2024~\citepalias{dcwp2024minpay}, and is calibrated to these disclosures and to public filings. None of its data is generated by a language model. We project this world into an Oracle EBS warehouse that omits the simulator's latent state, so tasks are harder to solve than to verify~\citep{song2025mind}. The agent must reconstruct facts from the warehouse, while the grader reads them off the state. Agents work in a sandboxed Python environment with statistics, machine-learning, and optimization libraries (Appendix~\ref{app:tools}), and file their decisions and results through a mission-control interface rather than returning a query.
Because the grader knows the latent state, it can grade a decision by its consequences. For instance, a list of banned accounts is scored by the fraud losses it prevents, including losses from fraud that the platform never detected, net of the revenue lost from wrongly banned customers.

Simulated environments are an established practice in a broad range of domains~\citep{altman2023realistic,lopez2016paysim,trivedi2024appworld,barres2025tau,xu2026theagentcompany,huang2025crmarena}, and synthetic warehouses have long been used to benchmark data systems~\citep{nambiar2006making,ghazal2013bigbench}. Even Spider 2.0 draws a fifth of its questions from synthetic, obfuscated, or textbook sample schemas (Appendix~\ref{app:spider2-provenance}). To our knowledge, no prior simulated benchmark combines an enterprise-scale warehouse whose tables must agree with one another and a grader that scores the consequences of the agent's actions against the simulator's latent state.

We evaluate 14 frontier and open-weight models on \ours. The strongest, Claude Opus 5.5, solves 34.8\% of tasks and averages 59.5 points. Nine of the fourteen average below 35. Models often analyze the wrong quantity or optimize the wrong objective.

We make the following contributions:
\begin{itemize}
  \item A synthetic, large-scale public ERP dataset in Oracle EBS format for a 2024 New York City food delivery platform, comprising 235 mutually constraining tables, 81 million orders, 3.4 million active customers, and 7.5 billion rows, in which an order resolves into dispatch decisions, courier pay, merchant payouts, and balanced general-ledger journals.
  \item A grader that scores the consequences of an action rather than the correctness of a query, using the simulator's latent state as ground truth, for instance, to score bans by the fraud losses they prevent and forecasts against held-out months, supporting tasks across fraud detection, forecasting, and financial planning.
  \item A benchmark of \ntasks such tasks, from publishing a dashboard data source to fitting forecasts and banning fraudulent accounts, with an evaluation of 14 frontier and open-weight models.
\end{itemize}

\ours is public. One world's warehouse is released on Hugging Face under CC BY 4.0,\footnote{\url{https://huggingface.co/datasets/textql/Argo-Bench}} and the tasks, reference agent, tool server, and sandboxes under the Apache License 2.0.\footnote{\url{https://github.com/TextQLLabs/Argo-Bench}} The numbers and experiments in this paper use a second world from a private seed, with different customers, couriers, and answer keys. The leaderboard is scored only on this world, so a score cannot be earned by memorizing the released warehouse. A demo at \url{https://argo-bench.com} allows readers to browse the orders and deliveries of the released world at a 10\% scale.

%% file: sec/02_benchmark.tex
\section{Benchmark Construction}
\label{sec:method}

\subsection{World Simulation}
\label{sec:environment}

We simulate a food delivery platform in New York City (NYC) in 2024, similar to DoorDash, Grubhub, or Uber Eats. Public data on such platforms is aggregate. The city's quarterly reports and the platforms' own filings give totals, but order-level records cannot be released without exposing the platform's customers, couriers, and margins. The world is therefore built from 34 public datasets and reports, each lending one mechanism (Appendix~\ref{app:donors}). Uber and Lyft trips, for instance, give the time to drive between two zones at a given hour, and MenuStat the menus for restaurant chains. Donors play one of three roles. \emph{Identity} donors are public records of real entities in the city, such as its 45,834 restaurants, 1.07 million addresses, and 260 taxi zones, and enter the world as they are, except that the released warehouse renames some restaurants (see the ethics statement). \emph{Shape} donors are measured elsewhere, on other people, in another city, or in another year, and lend the world only a distribution. \emph{Anchors} are published totals that the world is calibrated to reproduce but never samples records from.

\begin{figure}[t]
\centering
\includegraphics[width=\textwidth]{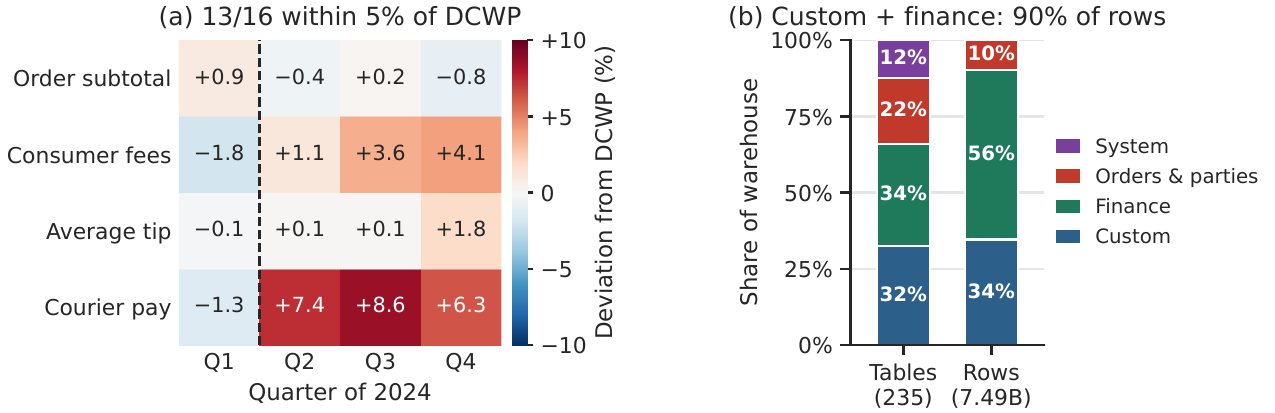}
\caption{The simulated world matches the city's reported economics, and its warehouse follows ERP conventions. (a) Deviation of per-delivery economics from the 2024 NYC DCWP quarterly anchors. Courier pay runs 6--9\% above the anchor after the April minimum-pay increase, and all other figures stay within 5\%. (b) Shares of the warehouse's 235 tables and 7.49 billion rows by EBS module group. Custom extensions alone hold about a third of both the tables and the rows.}
\label{fig:calibration}
\end{figure}

We chose food delivery in NYC because its economics are unusually well documented. Delivery apps must report their monthly orders, consumer spending, merchant fees, courier earnings, productivity, and hours worked to the NYC Department of Consumer and Worker Protection (DCWP), which publishes them quarterly~\citepalias{dcwp2024data}, and the 10-K filings of DoorDash and Grubhub give the shape of a platform's balance sheet~\citep{doordash2024tenk,doordash2025tenk,grubhub2021tenk}. Following \citet{walonoski2018synthea}, we calibrate the world to these anchors, sized as a dominant platform. The result has 81 million orders, about 55\% of the 148 million deliveries that apps reported to the DCWP for 2024, and 3.4 million active customers, and it must balance incentives on three sides: quests and suggested pay for couriers, promotions and surge pricing for customers, and co-funded campaigns for merchants. Its per-delivery economics stay within 5\% of the DCWP's figures in 13 of 16 quarterly comparisons (Figure~\ref{fig:calibration}a).

We selected 2024 because it contains a real extrinsic shock to these economics. NYC began enforcing a minimum pay rate of \$17.96 per hour before tips for app-based restaurant delivery workers in December 2023 and raised it to \$19.56 on April 1, 2024~\citepalias{dcwp2024minpay}. We model the platforms' response with a new courier scheduler that activates on that date, after which courier pay runs 6--9\% above the anchor (Figure~\ref{fig:calibration}a). The levers our platform uses may differ from those of the real platforms, but the aggregate effect is the same in direction and, to within 9\%, in size, and because the world absorbs the same shock, we can ask realistic forecasting questions about it (Appendix~\ref{app:minpay}).

Additionally, we insert fraud patterns that public evidence shows are major problems for delivery platforms. Couriers steal orders after pickup~\citep{aljazeera2025devagiri}, spoof their GPS~\citep{incognia2022spoofing}, grab offers with bots~\citep{chapman2020instacart}, or rent out their accounts~\citep{doordash2024sharing}. On the customer side, rings of new accounts farm promotions~\citep{doordash2025tenk,incognia2025gig}, and stolen cards fund account takeovers and bust-outs~\citep{doordash2025tenk,whittaker2018doordash}. Storefronts may be shells~\citep{heier2023ubereats} or collude with couriers or regular customers on refunds~\citep{doordash2023tenk}, and their payouts can be diverted to changed bank accounts~\citep{maycock2024wokhouse}. Each pattern is calibrated both to how separable real card fraud is and to how often honest customers share a device, an address, or a card, since the latter sets a detector's precision (Appendix~\ref{app:donors}). The simulator generates the world from the donors, calibrates it to the anchors, inserts the fraud, and projects the result to the EBS format (Section~\ref{sec:projection}).

\subsection{Warehouse Design and Validation}
\label{sec:projection}

The simulator's last step projects the world into what an analyst actually sees: the analytics export of a greenfield Oracle E-Business Suite (EBS) 12.2 instance. We chose EBS because its data model is publicly documented, so our schema can be verified against a reference. We designed the schema with three ERP consultants who have 14 to 31 years of experience.

Every standard table and column in our warehouse exists in the data dictionary of Oracle's EBS 12.2 Vision instance, the demo environment Oracle provides as a reference. Our tables, however, carry on average 52\% of the columns of their Vision counterparts. EBS serves every industry, and analytics exports omit the columns a business does not use, here unused flexfields (a third of the omitted columns) and features such as shipping, inventory, foreign currency, and withholding tax. These columns would be empty in this business's data, so no task loses information by their omission.

The warehouse does not contain data drift or inconsistencies, such as deprecated tables that overlap active ones or figures that fail to reconcile across tables. These inconsistencies sometimes accumulate in real data warehouses over years of migrations and acquisitions~\citep{vogelsgesang2018get}. Though the consultants named this the most significant difference from their customers' systems, we deliberately chose to keep this discrepancy. By doing so, we keep the ground truth unambiguous: if a legacy table were to disagree with an active one, the correct answer would depend on undocumented conventions. Mature warehouses compensate for their idiosyncrasies with semantic layers, data models, and institutional knowledge~\citep{kandel2012enterprise}. While we could provide such a layer alongside a more realistic messy warehouse, this would shift the evaluation's focus to testing an ability to use a curated layer. We instead aim to test an understanding of enterprise data organization: production workloads reuse only a few dozen combinations of hundreds of tables ~\citep{van2024tpc}, and an agent new to a warehouse must discover which ones matter by deciding what and how much of the warehouse to explore. A greenfield warehouse isolates this skill of understanding and exploring enterprise data organization.

Otherwise, the consultants found the schema realistic, with two further omissions. It records no foreign-currency transactions, since no donor dataset covers the currencies visitors pay in, and it has only 43 balance sheet accounts (Section~\ref{sec:limitations}).

\subsection{Task Design}
\label{sec:tasks}

\begin{figure}[t]
\centering
\includegraphics[width=\textwidth]{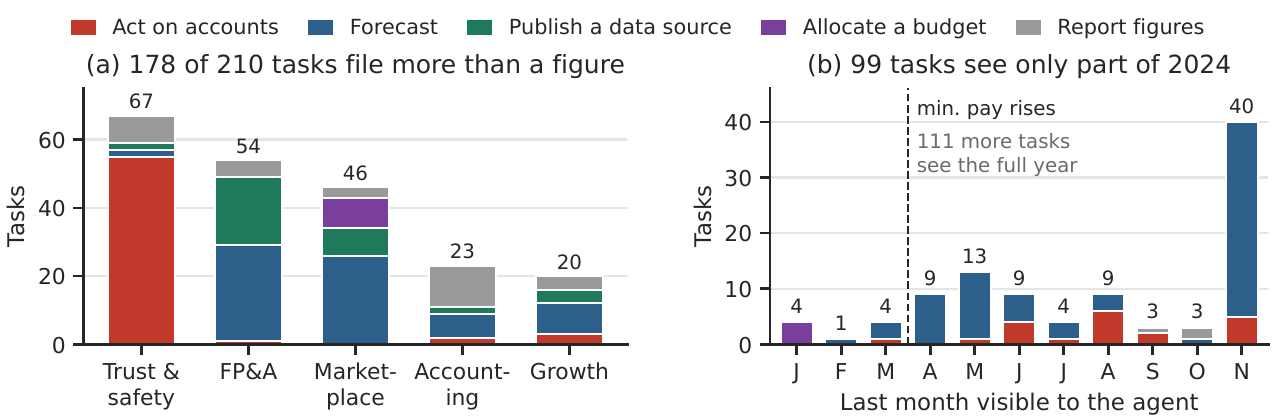}
\caption{Tasks in each business area vary by expected final action and the snapshot of data given. (a) Tasks per business area, grouped by action taken. Multi-part tasks are counted by their first filing. (b) The last month of 2024 visible to the agent in the 99 tasks with a partial view. Partial views are used mostly in forecasting tasks, which are graded against the months after the cutoff.}
\label{fig:tasks}
\end{figure}

\ours contains \ntasks tasks in five business areas (Figure~\ref{fig:tasks}a). Trust and safety tasks are enforcement, where the agent finds fraud and abuse and acts on the accounts involved. FP\&A tasks forecast unit economics and rebuild finance dashboards, marketplace tasks forecast courier supply and allocate budgets such as courier bonuses, accounting tasks report final values after the fact, and growth tasks measure, forecast, and publish the results of promotions and memberships.

Each task has four parts (Appendix~\ref{app:grading}). The \emph{prompt} states the problem as a stakeholder would. The \emph{scope} sets the last month of 2024 visible to the agent. The \emph{expectations} list the filings the grader requires, each with its action, keys, and grading mode. The \emph{answer key} is frozen before any run and comes from SQL over the latent tables or from the simulator's own labels, such as which couriers stole orders. Because the world is simulated, these labels are exact and need no anonymization~\citep{gadotti2024anonymization}.

Prompts cover a range of writing styles and levels of detail. Some reference the grading criteria or the exact output expected, while others are more subtle. This reflects how real users pose data questions: loosely, as high-level business questions, and in no set style or template~\citep{kandel2012enterprise}. It also tests the skill of exploring data organization, since a less detailed prompt leaves the agent to discover which tables and conventions the question depends on. Several scenarios come in variants that differ only in such detail, and Appendix~\ref{app:findings} compares them.

Unlike most data science and analytics benchmarks, \ours does not ask agents to return a query. Agents file actions to a mission-control interface through a Python library in their sandbox (Appendix~\ref{app:prompt}). This design has four advantages. First, it permits advanced data science tasks which require machine learning, operations research, and mathematical optimization libraries to complete. Second, it supports end-to-end workflows. Emitting the correct SQL is not enough in practice, as real tasks require taking actions, e.g., rebalancing courier incentives across zones and hours, banning a set of users who are likely committing fraud, or holding the payouts of a suspicious merchant.
Third, filings are explicit declarations of intent. When benchmarks compare SQL results~\citep{li2023can,lei2025spider,chen2024beaver}, it is hard to determine whether a close number is the agent's answer or an intermediate result, whereas a filing states the value, interval, or reason the agent commits to, which also makes partial credit well defined. Fourth, grading is objective, unlike the LLM-as-a-judge evaluation~\citep{zheng2023judging} used in many related works.

Of the \ntasks tasks, \nacted act on accounts, file a forecast, allocate a budget, or publish a dashboard data source, and \npartial see only up to a cutoff month, as an analyst would at that date, so forecasts are graded on months the agent has not seen (Figure~\ref{fig:tasks}). Prompts average \npromptwords words, 51 tasks require more than one filing, and the reference solution in Figure~\ref{fig:overview} joins six tables across three EBS modules to recover the minimum-pay rule and fit an interval. In size, \ours matches long-horizon agent benchmarks such as TheAgentCompany (175 tasks)~\citep{xu2026theagentcompany}, $\tau$-bench (165)~\citep{yao2024tau}, KramaBench (104)~\citep{lai2026kramabench}, and ELT-Bench (100)~\citep{jin2025elt}.

Each expectation is scored from 0 to 100 (a forecast from $-200$) by one of nine grading modes (Appendix~\ref{app:grading}). Forecasts are scored by their weighted interval score~\citep{bracher2021evaluating} on a scale set by a reference forecast fixed before the outcome, so that filing one's true median and interval is the best strategy, ban lists by the cost they save relative to banning nobody or everybody~\citep{elkan2001foundations}, budget allocations by the share of the attainable savings that the simulator realizes, and data sources and reported figures by their values. A task's score is the mean of its expectations' scores, weighted as the task specifies. A run that files nothing where the key expects action scores zero on every expectation ($-200$ on a forecast, the lowest a forecast can score).

We release the warehouse of one world on Hugging Face and keep a second, generated from a private seed, for official grading. We ran our experiments on BigQuery (941~GB uncompressed), and the released warehouse is a set of 1,219 Parquet files (76.5~GB), with a dataset card giving setup instructions for BigQuery, DuckDB, Snowflake, Trino, Delta Lake, and Iceberg. The two seeds share the simulator and its calibration, but every ID, customer, restaurant, and courier differs.\footnote{Because the economy depends on the seed, volumes also differ slightly. The public warehouse has 7.54 billion rows compared to the 7.49 billion in the private world described in this paper.} Table~\ref{tab:comparison} compares \ours with prior benchmarks.

\begin{table}[t]
\caption{\ours pairs an ERP-scale warehouse with a Python sandbox for machine learning and optimization, and grades an agent's actions against a latent state. Tables and rows are per database. Entries marked n/r are not reported by the benchmark. A coherent system is a single enterprise application whose tables must reconcile. Actions are graded when the benchmark scores what the agent changes or files rather than an answer it returns.}
\label{tab:comparison}
\centering
\footnotesize
\setlength{\tabcolsep}{2.6pt}
\begin{tabular}{@{}lrrlcccl@{}}
\toprule
Benchmark & \makecell{Tables\\per DB} & \makecell{Rows\\per DB} & Data & \makecell{Coherent\\system} & \makecell{Python\\and ML} & \makecell{Actions\\graded} & Ground truth \\
\midrule
BIRD~\citep{li2023can} & 7.3 & 549K & Public & \xmark & \xmark & \xmark & Gold SQL \\
Spider 2.0~\citep{lei2025spider} & 52.6\rlap{$^\dagger$} & n/r & Public & \xmark & \xmark & \xmark & Gold SQL \\
BEAVER~\citep{chen2024beaver} & 101.5 & n/r & Private & \cmark & \xmark & \xmark & Logged SQL \\
DSBench~\citep{jing2025dsbench} & n/r & n/r & Public & \xmark & \cmark & \xmark & Answer keys \\
DABstep~\citep{egg2025dabstep} & n/r & 138K & Real & \xmark & \cmark & \xmark & Answer keys \\
$\tau$-bench~\citep{yao2024tau} & 3 & 2.8K & LLM-made & \xmark & \xmark & \cmark & Goal state \\
CRMArena-Pro~\citep{huang2025crmarena} & 25 & 55K & LLM-made & \cmark & \xmark & \xmark & Generator \\
\midrule
\rowcolor{oursrow}\ours & \textbf{235} & \textbf{7.49B} & Simulated & \cmark & \cmark & \cmark & Latent state \\
\bottomrule
\end{tabular}
\par\smallskip
\parbox{\linewidth}{\scriptsize $^\dagger$Spider 2.0-Lite, as computed by \citet{chen2024beaver}.}
\end{table}

%% file: sec/03_evaluation.tex
\section{Evaluation}
\label{sec:evaluation}

We run all experiments on Inspect AI 0.3.263~\citep{inspectai}, an open-source evaluation framework, on Google Kubernetes Engine. Each agent works in its own gVisor~\citep{young2019true} sandbox with 25 preinstalled Python libraries, an empty file system, and no outbound internet access, and files to its own mission-control instance through a Python library (\texttt{mission\_control.py}) that emulates a company's internal tooling. A run ends after 500 model turns, to stop models that loop without progress, and Appendix~\ref{app:tools} lists the other limits. We evaluate models available in September 2026 across price ranges (Table~\ref{tab:results}), calling open-weight models through Fireworks serverless endpoints and the others through their developers' APIs, with default sampling settings for every model.

\begin{table}[t]
\caption{Results on \ours{} (210 tasks), each model at the highest reasoning effort we ran, extra-high where offered (Appendix~\ref{app:findings} gives every level). Solved is the share of tasks scoring at least 95, and score is the mean task score out of 100, with a model's forecast tasks floored at 0 as a block (Appendix~\ref{app:grading}). Domain columns give mean scores. Steps (model calls) and cost (API spend, excluding BigQuery) are per-task averages. Best results are in \textbf{bold}.}
\label{tab:results}
\centering
\small
\setlength{\tabcolsep}{4pt}
\begin{tabular}{@{}lcccccccrr@{}}%
\toprule
 & \multicolumn{2}{c}{Overall} & \multicolumn{5}{c}{Score by domain} & & \\
\cmidrule(lr){2-3}\cmidrule(lr){4-8}
Model & Solved (\%) & Score & Fcst. & Fraud & Fin. & Dash. & Comp. & Steps & Cost (\$) \\
\midrule
\multicolumn{10}{@{}l}{\textit{Proprietary}} \\
GPT-6 Astra       & 27.6 & 51.8 & \textbf{39.5} & 48.0 & 85.8 & 71.0 & 85.7 & 23 & 2.71 \\
GPT-6.1 Sol       & 24.8 & 49.5 & 38.4 & 47.6 & 87.1 & 63.4 & 82.8 & 24 & 0.49 \\
GPT-6 Sol         & 17.6 & 36.8 & 27.9 & 35.4 & 75.8 & 37.3 & 75.7 & 38 & 0.85 \\
GPT-6 Luna        & 7.6 & 15.0 & 2.1 & 19.1 & 56.1 & 10.1 & 42.9 & 39 & \textbf{0.06} \\
Claude Opus 5.5   & \textbf{34.8} & \textbf{59.5} & 35.3 & \textbf{66.1} & \textbf{90.0} & \textbf{84.6} & 81.6 & 82 & 4.71 \\
Claude Sonnet 5.5 & 28.6 & 51.8 & 34.6 & 62.5 & 75.6 & 54.1 & \textbf{86.9} & 77 & 3.74 \\
Claude Sonnet 5   & 9.5 & 17.3 & 2.5 & 23.7 & 57.0 & 21.5 & 23.8 & 55 & 2.08 \\
Claude Haiku 4.5  & 1.4 & 5.5 & 0.4 & 9.3 & 19.2 & 0.0 & 35.7 & 31 & 0.20 \\
Gemini 3.8 Flash  & 16.7 & 26.2 & 3.8 & 33.4 & 61.3 & 36.5 & 85.7 & 197 & 3.94 \\
Muse Spark 1.3    & 13.8 & 20.7 & 3.8 & 24.0 & 46.8 & 26.3 & 85.7 & 250 & 4.89 \\
\midrule
\multicolumn{10}{@{}l}{\textit{Open-weight}} \\
Kimi K3           & 14.3 & 28.4 & 15.8 & 33.6 & 73.3 & 18.8 & 81.0 & 85 & 4.52 \\
GLM 5.3 Flash     & 12.4 & 21.2 & 3.2 & 29.8 & 57.5 & 23.0 & 61.9 & 92 & 0.35 \\
DeepSeek V4.1 Flash & 17.6 & 25.4 & 4.6 & 30.8 & 56.9 & 41.7 & 75.0 & 125 & 0.42 \\
Qwen 3.8 Max      & 17.1 & 24.4 & 6.1 & 24.5 & 66.1 & 35.5 & 65.5 & 77 & 2.72 \\
\bottomrule
\end{tabular}
\end{table}

\subsection{Results}
\label{sec:results}

Claude Opus 5.5 leads overall and in three of the five domains (Table~\ref{tab:results}), GPT-6 Astra leads in forecasting, and Claude Sonnet 5.5 leads in compliance. More reasoning effort helps the GPT-6 and Claude models at every step, with diminishing returns for Opus, which gains 17.0 points from low to medium effort, 5.9 from medium to high, and 4.0 from high to extra-high, while Claude Sonnet 5.5 gains 14.9 and 11.5 over the last two steps. Gemini 3.8 Flash gains 8.6 points from low to medium [4.9, 16.3] and Qwen 3.8 Max gains 5.2, and neither gains detectably afterward. Muse Spark 1.3 changes by less than 2 points past medium, and DeepSeek V4.1 Flash moves only at extra-high (Appendix~\ref{app:findings}). Gemini 3.8 Flash and Muse Spark 1.3 also average 197 and 250 model calls per task, compared to 82 for Opus, without scoring higher, and 57\% of Muse's warehouse spend goes to tasks on which it scores below 5 out of 100 (Appendix~\ref{app:bigquery}). 

Many tasks are prompt variants of one scenario, so we resample the 146 base scenarios when bootstrapping. The resulting 95\% confidence intervals are about $\pm$2 to 8 points on the score and up to $\pm$8 points on the solved rate. Paired by task, Opus leads each of the next three models (GPT-6 Astra, Claude Sonnet 5.5, and GPT-6.1 Sol) by 7.7 to 10.0 points, and those three are not separated from one another. GPT-6.1 Sol leads its predecessor GPT-6 Sol by 12.8 points [6.8, 19.1], GPT-6 Sol leads Kimi K3 by 8.4 [0.5, 15.4], and the intervals of the next five overlap (Appendix~\ref{app:findings}).

\subsection{Findings}
\label{sec:findings}

Many failing runs use sound methods but read the wrong record, optimize the wrong objective, or measure the wrong quantity. Passing runs check definitions against a second source (Appendix~\ref{app:findings}).

\paragraph{Wrong record.} A marketing dashboard asks whether each discount offer paid off against the same push notification with the coupon left out for every thousand customers it was sent to, for members and non-members. Only Claude Opus 5.5, GPT-6 Astra, and GPT-6.1 Sol at extra-high effort publish the right table. What separates them is who counted as a member at the moment of targeting. A member whose card is declined keeps the benefits for seven days of grace and then loses them, but the contract stays on the books until it is canceled weeks later. Twenty-four of the failing runs take membership from the contract's dates, and seven of them get every other figure right to within a few dollars. The three passing runs replay the billing history instead and check the result against the orders on which a member benefit was actually applied. Opus reads the seven-day grace off those orders, and GPT-6.1 Sol starts from the contract dates, finds 1,057 orders that its rule calls members' and that received no benefit, and starts over.

\paragraph{Wrong objective.} One family of tasks asks where to cut \$9.6 million from the annual budget for courier bonuses (quests). At extra-high effort and without hints, GPT-6 Astra finds the zones and hours where quests were randomly withheld, estimates supply responses with fixed effects and partial pooling, and cuts where quests buy the fewest courier-hours. The platform pays for quests to avoid surge pay; however, under the simulator's response model, the plan loses \$86{,}281 where a uniform cut would save \$0.40 million, and it scores 0. From the same prompt, Claude Opus 5.5 at extra-high effort finds that quests substitute for surge pay, cuts where they save the least surge per bonus dollar, and saves \$3.09 million of an attainable \$3.12 million (score 99). Stating what quests are for and that a holdout exists raises the mean over all settings from 18.5 to 61.7.

\paragraph{Wrong quantity.} Many failures come from measuring a quantity other than the one the prompt asks for. Of the 47 completed runs of a task that sizes the courier location service, 46 miss all 12 monthly counts despite a 1\% tolerance because the warehouse keeps only hourly idle check-ins while the app sends them every half hour. GPT-6 Astra notices the gap, writes that its count does not establish how many reports the app sent, and files it anyway. Only Claude Sonnet 5.5 at extra-high effort adds the missing half-hourly check-ins back. Across the dashboard tasks, 66.9\% of the data sources that pass their structural contract score zero on their values.

\paragraph{Overconfident forecasts.} Across 4,553 forecast series from 3,346 runs on 72 tasks, nominal 80\% intervals contain the realized value only 44.8\% of the time. Apart from its floor, the grade is proper, so this overconfidence costs models points in expectation, although a per-series skill score clipped at zero would have rewarded a narrower interval in 72\% of series (Appendix~\ref{app:findings}). Grades also depend on the reference, which sets each series' scale. Against the tighter reference of their harder variant, the June base-pay forecasts fall from a mean grade of 84.7 to 6.0, although their median absolute error is only 1.57\%. Appendix~\ref{app:findings} reports coverage and absolute error.

%% file: sec/04_related_work.tex
\section{Related Work}
\label{sec:related}

\paragraph{Text-to-SQL and data science benchmarks.} Text-to-SQL benchmarks have moved from databases with a handful of tables each~\citep{yu2018spider,li2023can} to enterprise-scale schemas~\citep{lei2025spider} and private data warehouses~\citep{chen2024beaver}, and data science benchmarks extend evaluation to multi-step analysis, data lakes, and data pipelines~\citep{jing2025dsbench,egg2025dabstep,lai2026kramabench,lei2026dacomp,jin2025elt}. These benchmarks compare an agent's output to a gold answer, which audits have found to be frequently wrong~\citep{wretblad2024understanding,jin2026pervasive}, or to expert conclusions, often scored by an LLM judge~\citep{gu2024blade,sahu2025insightbench}. \ours instead grades the actions that an agent files against the simulator's latent state.

\paragraph{Agent benchmarks in simulated environments.} Simulated environments are the standard way to evaluate agents that act. AppWorld~\citep{trivedi2024appworld} and $\tau$-bench~\citep{yao2024tau,barres2025tau} check the final state of the environment's database. TheAgentCompany~\citep{xu2026theagentcompany} scores checkpoints in a simulated software company, and CRMArena-Pro~\citep{huang2025crmarena} shapes LLM-generated CRM records with latent variables. Vending-Bench and Business Arena score the net worth of a business that the agent runs~\citep{backlund2025vending,pan2026business}. In these benchmarks, the state that determines success is either observable to the agent or changed by the agent inside a stylized game. In \ours, it is withheld and must be reconstructed from an enterprise warehouse. Simulators calibrated to public statistics have likewise supplied ground truth that real data lacks for patient records~\citep{walonoski2018synthea} and money laundering~\citep{altman2023realistic}.

\paragraph{Generated enterprises with hidden ground truth.} AvalancheBench~\citep{kleczek2026avalanchebench} uses an LLM judge to score how much of a small latent e-commerce world an agent's report recovers. The Era by Eon benchmark~\citep{gruenbaum2026era} serves a generated company through a fleet of 66 simulated products, calibrated to operational and published statistics, and plants the records that answer each of its read-only questions, together with near misses. Two unrelated benchmarks named ERPBench evaluate decisions in a simulated manufacturer~\citep{zhang2026erpbench} and computer-use tasks in a live ERP system~\citep{bhagtani2026erpbench}. In contrast, the latent state of \ours is produced by the simulation itself rather than being planted. Its evidence is spread across an enterprise warehouse of 7.49 billion rows, and agents are graded on the consequences of the actions they file rather than on the answers they give.

%% file: sec/05_limitations.tex
\section{Limitations \& Future Work}
\label{sec:limitations}

The dataset still differs from the most complex ERP deployments in four ways. First, the dataset covers a single city, so it has no foreign currency and none of its representations as transaction, local, and reporting currency. Second, many complex warehouses combine several businesses, such as food delivery and grocery delivery, with shared accounts, such as driver payables and customer credits, mixing them in a single balance. Third, only one year is simulated. A longer history, such as 2017 to 2026, would add the market shocks of 2020 to 2022 to operations and forecasts. Fourth, only one ERP format is supported, and a natural extension would be to add SAP S/4HANA.

The simulator validates 23 distinct metrics, and we keep behaviors that public figures do not constrain out of scope for tasks. The in-world membership program in particular rests on weakly grounded assumptions. More generally, a simulator encodes the assumptions of its authors, and a generator that shares the simplifying assumptions of the systems under evaluation can make tasks easier than their real counterparts~\citep{leis2015good}. Calibration to aggregate targets also does not guarantee realistic tails~\citep{chen2019validity}. \ours therefore compares data agents and does not estimate their performance on a real company's warehouse.

Each setting of the model and effort has one run per task, and all tasks share one simulated world, so our confidence intervals reflect the choice of tasks rather than run-to-run variation. Scores on some tasks are also sensitive to prompt wording, so Appendix~\ref{app:findings} reports every hinted and unhinted pair, including one prompt we judged to be underspecified. Plan tasks are graded under the simulator's frozen response model, and Appendix~\ref{app:audit} lists open issues by task.

%% file: sec/06_conclusion.tex
\section{Conclusion}
\label{sec:conclusion}

We introduced \ours, which evaluates data agents on a simulated food delivery platform exported to an Oracle E-Business Suite warehouse of 235 tables and 7.49 billion rows. Because the simulator's latent state is withheld from the warehouse, \ours grades the facts agents reconstruct and the actions they file, with every answer key computed from ground truth. The strongest of 14 models solves 34.8\% of tasks and averages 59.5 points, and its most instructive failures are careful analyses of the wrong quantity, toward the wrong objective, or read from the wrong record. We release the public seed's warehouse, tasks, reference solutions, and harness, and hope that \ours helps measure progress toward data agents that can understand, navigate, and act within real enterprise data environments.

%% file: sec/07_statements.tex
\subsection*{AI use statement}
Large language models were used in three ways. First, as coding assistants for the simulator, graders, reference solutions, and figure and table code. Second, in the simulator's data work: tuning its parameters, cleaning donor datasets, and reconciling restaurant records across sources (Section~\ref{sec:environment}). No record in the simulated world is generated by a language model, and every value in the warehouse comes from the simulator, except the fictional names of the storefronts renamed in the released warehouse, which a language model drafted (see the ethics statement). Third, in writing: some task prompts were drafted by a language model and then curated and rewritten by the authors, and language models edited the text of the paper and checked its citations. The authors checked every claim, number, and citation, and take full responsibility for the content of the paper.

\subsection*{Ethics statement}
\ours contains no data about real people. Consumer and courier names are drawn from public name-frequency tables and assigned to simulated people, and every order, shift, and payment is simulated. Restaurants are real New York City businesses. Their names and addresses come from Overture Maps places matched to the city's inspection records, with some names updated to the business's current listing, and some storefronts the simulator opens during the year, such as relaunches and virtual brands, take the name of a real business that was not trading at the time. All behavior attributed to them in the world, including the fraud patterns of Section~\ref{sec:environment}, is simulated. The simulator draws which storefronts play a fraud role, not from any record of a business's conduct, so a label says nothing about the real business. Even so, in the released warehouse every storefront that plays a fraud role in any scenario carries a fictional name instead of its real one, and fields derived from the name, such as its contact email, follow the new name. The released questions and agent transcripts use the same fictional names. A language model drafted the fictional names to read like real New York restaurant names, so that a renamed storefront does not stand out and point to the answers, and each was checked against the city's inspection records and Overture Maps so that none is the name of a real restaurant. Addresses are unchanged, since the simulated geography depends on them. The donor datasets are used under their licenses. Those released for research or non-commercial use, such as the Yelp Open Dataset and the Grubhub MDRP instances, serve only as shape donors, and none of their records enters the world or the warehouse (Appendix~\ref{app:donors}). The fraud tasks reward detecting common schemes, not carrying them out. During development, one model escaped an insufficiently isolated sandbox and read the grader code (Appendix~\ref{app:reward-hacking}). The final runs use isolated sandboxes without internet access, and we report the incident so that others building agent benchmarks can guard against it.

\subsection*{Reproducibility statement}
Two artifacts are released. The warehouse of one seed is released on Hugging Face under CC BY 4.0 at \url{https://huggingface.co/datasets/textql/Argo-Bench}. The code that reproduces the paper's runs is released under the Apache License 2.0 at \url{https://github.com/TextQLLabs/Argo-Bench}. It contains the \ntasks questions, the reference agent on Inspect AI 0.3.263, its warehouse and Python tool servers, the mission-control console through which the agent files, the three sandboxes in which \texttt{run\_python} executes (a macOS Seatbelt profile, a Docker image with pinned libraries, and the network-less gVisor pod used for the paper's runs), loaders for DuckDB and BigQuery, and the model and reasoning-effort configuration for every rung reported. The system prompt and tools are given in Appendices~\ref{app:prompt} and~\ref{app:tools}, and Appendix~\ref{app:examples} shows three reference solutions in full.

The simulator and the graders are not released to prevent direct answer memorization. A run made with the released code exports a submission file recording every filing, the model and its settings, token usage, and how each run ended, which we score on a best-effort basis. The paper's runs queried a second world generated by the same simulator from a different seed (Section~\ref{sec:tasks}). The 15 questions that name specific couriers, storefronts, or promotion codes draw them from the public world by the same selection rule. Re-running the released code therefore reproduces the paper's procedure on a sibling world rather than its exact numbers. Proprietary models were accessed through their providers' APIs, and all other models through Fireworks serverless endpoints with default settings in September 2026. Results from proprietary APIs may drift as providers update their models.

%% file: sec/08_acknowledgments.tex
\subsubsection*{Acknowledgments}
We thank Angela Peng, Alexander Baumstark, and Ben Van Sleen for their work on the design of the benchmark and on the evaluations, and JS Irick, David Dixon, and Scott Cairncross for building the ERP, FP\&A, and reporting components of the simulator. We also thank Mark Hay, Ben Mains, Matthew Abate, Sergi Domingo, and our colleagues at TextQL for their feedback and support, the New York City agencies whose public reports and records the simulator is built on, and the maintainers of Inspect AI.

%% file: sec/appendix/A_donor_datasets.tex
\section{Donor datasets}
\label{app:donors}
Table~\ref{tab:donors} lists 34 public datasets and reports that were used to  build the world (Section~\ref{sec:environment}), categorized by the part of the world they support. Donor datasets released for research or non-commercial use, among them the Yelp Open Dataset~\citep{yelpopen} and the Google Local reviews~\citep{li2022uctopic,yan2023personalized}, serve only as shape donors: none of their records reach the world or the released warehouse, only parameters fitted on them.

\begin{table}[!ht]
\caption{The donor datasets, by the part of the world they build. Role: I = identity (real NYC records, used as they are), S = shape (a distribution measured elsewhere and transferred), A = anchor (published figures the world is calibrated to). $\dagger$~Released for research or non-commercial use, and used only as a shape donor. $\ddagger$~Used for validation only: no fit reads it.}
\label{tab:donors}
\centering
\footnotesize\renewcommand{\arraystretch}{0.9}
\begin{tabularx}{\linewidth}{@{}>{\raggedright\arraybackslash}p{4.2cm}>{\raggedright\arraybackslash}X c@{}}
\toprule
Donor & Used for & Role \\
\midrule
\multicolumn{3}{@{}l}{\textit{Geography}} \\
Overture Places & The 45,834 restaurants, with names and locations & I \\
DOHMH inspections & Cuisine, license, grades, openings and closures since 2017 & I \\
Overture Addresses & 1.07M residential address points & I \\
MapPLUTO & Building use, units, and floor area of 857k tax lots & I \\
TLC taxi zones & The 260 zones that partition the city & I \\
OpenStreetMap & Street network and bridge access & I \\
ACS 2020--2024 & Tract composition, households, and zone income & I \\
\midrule
\multicolumn{3}{@{}l}{\textit{Customers}} \\
2020 Census surnames & Surnames by origin & S \\
WGND 2.0 & Given names by country and gender & S \\
NYC baby names & Given names by NYC community & S \\
SSA New York births & Given names by birth year, for adults in 2024 & S \\
NANPA code assignments & Phone exchanges assigned in the NYC area & I \\
LODES 2023 & Home-to-work flows for office anchors & S \\
\midrule
\multicolumn{3}{@{}l}{\textit{Demand timing}} \\
TLC high-volume FHV trips & Demand by zone, hour, and weekday & S \\
GHCN-Daily Central Park & Daily rain, snow, and temperature & I \\
NOAA LCD & Hourly precipitation & I \\
DoorDash 2015 export & Event-day demand shapes & S \\
ATUS 2024 & How often and when people host guests & S \\
\midrule
\multicolumn{3}{@{}l}{\textit{Menus and prices}} \\
Google Local reviews$^\dagger$ & Storefront popularity and quality rank & S \\
Yelp Open Dataset$^\dagger$ & Popularity curve and check-in tempo by cuisine & S \\
MenuStat & Chain menus & I \\
Wayback Machine offers & 1,648 promotions: values, limits, durations & S \\
\midrule
\multicolumn{3}{@{}l}{\textit{Kitchens}} \\
Grubhub MDRP$^\dagger$ & Preparation times at named restaurants & S \\
QSR drive-thru study & Lower tail of preparation times & A \\
NYC Open Restaurants & Seating, which bounds kitchen load & I \\
\midrule
\multicolumn{3}{@{}l}{\textit{Couriers}} \\
Chicago taxi trips & Shift lengths and trips per shift & S \\
NYC Citi Bike & E-bike speeds by street class & S \\
NYC yellow cabs, 2013--2015 & Stop locations within a zone & S \\
DCWP worker study 2022 & Vehicle mix and connected time & A \\
\midrule
\multicolumn{3}{@{}l}{\textit{Fraud and accounting}} \\
IEEE-CIS$^\dagger$ & Fraud separability and reuse; email domains & S \\
RBA logins & How often honest users share devices and networks & S \\
UCI Online Retail II & Returns and partial refunds & S \\
DCWP quarterly reports & Orders, fees, and courier pay & A \\
SEC 10-K filings & Balance-sheet and expense shapes & A \\
\bottomrule
\end{tabularx}
\end{table}

%% file: sec/appendix/B_minimum_pay.tex
\section{Questions enabled by the minimum-pay change}
\label{app:minpay}

On April 1, 2024, NYC raised the minimum pay rate for app-based restaurant delivery workers from \$17.96 to \$19.56 per hour~\citepalias{dcwp2024minpay}, and the simulated platform responds on the same day. Its dispatcher starts restricting when couriers can go online, and its per-delivery pay card drops (Section~\ref{sec:environment}). Whenever a week's pay falls short of the minimum, the platform tops it up with a true-up on the weekly courier pay invoice. The shock breaks the stationarity that simple forecasts rely on, and it changes how several standard figures must be computed. Table~\ref{tab:minpay} groups the tasks that depend on it.

\begin{table}[!ht]
\caption{Tasks that depend on the April 2024 minimum-pay change. Cutoff is the last month of 2024 visible to the agent.}
\label{tab:minpay}
\centering
\small
\begin{tabularx}{\linewidth}{@{}>{\raggedright\arraybackslash}p{1.9cm}>{\raggedright\arraybackslash}X>{\raggedright\arraybackslash}p{4.15cm}>{\raggedright\arraybackslash}p{1.75cm}@{}}
\toprule
Category & Task & Examples & Cutoff \\
\midrule
Before the shock & Forecast April's true-ups and hours from the first quarter, before any true-up has been paid & \texttt{fc-08-true-ups-march}, \texttt{fc-13-april} & March \\
Just after & Recover the true-up rule from the first weeks under it and forecast pay, hours, true-ups or margin for May to the third quarter & \texttt{fc-08}, \texttt{fc-10}, \texttt{fc-11}, \texttt{fc-12} (Table~\ref{tab:example-forecast}) & April or May \\
Later in the year & Forecast July's true-ups, third-quarter courier costs, and December's no-courier cancellations through further dispatcher and pay-card changes & \texttt{fc-10t2}, \texttt{fc-13-july}, \texttt{fc-19}, \texttt{fc-16} & June to November \\
Regulatory checks & Find the pay periods that paid couriers short, restate pay per connected hour on the rule's basis, and size back pay after a complaint & \texttt{pvl-01}, \texttt{vd-e3}, \texttt{fc-14-back-pay} & August to December \\
Dashboards & Rebuild margin and courier-cost tiles whose true-up component starts in April & \texttt{ds-22} to \texttt{ds-25}, \texttt{ds-33}, \texttt{ds-36t}, \texttt{erp-01} & December \\
\bottomrule
\end{tabularx}
\end{table}

%% file: sec/appendix/C_warehouse_tables.tex
\section{Warehouse tables}
\label{app:tables}

Table~\ref{tab:tables} lists every table in the warehouse with its column and row counts, grouped by the Oracle E-Business Suite module that owns it. The 159 standard tables fall under 15 Oracle products. The 76 custom extensions are grouped by the part of the business they record.

\input{sec/appendix/C_warehouse_tables_list}

%% file: sec/appendix/C_warehouse_tables_list.tex
\newcommand{\tblfont}{\fontsize{6.6}{7.9}\selectfont}
\newcommand{\tbltt}{\fontfamily{lmtt}\selectfont}
\newlength{\tblnum}\setlength{\tblnum}{15pt}
\newlength{\tblrows}\setlength{\tblrows}{27pt}
\newlength{\tblbarw}\setlength{\tblbarw}{17pt}
\newcolumntype{T}{%
  >{\leavevmode\tbltt\color{figink}}p{\dimexpr\linewidth-\tblnum-\tblrows-\tblbarw-6pt\relax}@{}%
  >{\raggedleft\arraybackslash\leavevmode\color{figink2}}p{\tblnum}@{}%
  >{\raggedleft\arraybackslash}p{\tblrows}@{\hspace{6pt}}%
  p{\tblbarw}}
\newcommand{\tblbar}[2]{\textcolor{#1!55}{\rule[0.1ex]{#2}{1.1ex}}}
\newcommand{\tblspan}[1]{\multicolumn{4}{@{}p{\linewidth}@{}}{#1}}
\newcommand{\tblcolhead}{%
  \multicolumn{1}{@{}l}{\color{figgreytag}\itshape Table} &
  \multicolumn{1}{r@{}}{\color{figgreytag}\itshape Cols} &
  \multicolumn{1}{r@{\hspace{6pt}}}{\color{figgreytag}\itshape Rows} & \\[-0.3ex]}
\newcommand{\tblfamily}[4]{%
  \tblspan{\rule{0pt}{3.4ex}\color{#1!80!black}%
    {\fontsize{8}{9}\selectfont\bfseries #2}\hfill
    {\fontsize{6.6}{7.9}\selectfont #3 tables\enspace\textperiodcentered\enspace #4 rows}}\\[0.2ex]
  \arrayrulecolor{#1}\specialrule{0.9pt}{0pt}{0.25ex}\arrayrulecolor{black}}
\newcommand{\tblmodule}[5]{%
  \ifnum#5=1 \noalign{\vskip 0.5ex}\fi
  \rowcolor{#1!12}\multicolumn{4}{@{\hspace{2.5pt}}p{\dimexpr\linewidth-5pt\relax}@{\hspace{2.5pt}}}{%
    \rule{0pt}{2.25ex}{\color{figink}\bfseries #2}\hfill
    {\color{figink2}#3 tables\enspace\textperiodcentered\enspace #4 rows}}\\[0.1ex]}
\newcommand{\tblrow}[4]{#1 & #2 & #3 & #4\\}

\begin{table}[p]
\tblfont\setlength{\tabcolsep}{2.2pt}\renewcommand{\arraystretch}{1}
\caption{The warehouse's 235 tables{,} 3{,}949 columns and 7{,}491{,}508{,}904 rows, by Oracle E-Business Suite module. Families and colors follow Figure~\ref{fig:calibration}b. Bars show rows on a log scale. The six empty tables are four interface tables, which a completed import leaves empty, \texttt{AR\_ADJUSTMENTS\_ALL}, and \texttt{HZ\_CUST\_ACCOUNT\_ROLES}.}
\label{tab:tables}\vspace{-0.6ex}
\centering
\begin{minipage}[t]{0.475\linewidth}
\begin{tabular}[t]{@{}T@{}}
\tblfamily{figorange}{Orders \& parties}{51}{723\,M}
\tblcolhead
\tblmodule{figorange}{Order Management}{6}{580\,M}{0}
\tblrow{OE\_ORDER\_HEADERS\_ALL}{41}{81\,M}{\tblbar{figorange}{11.07pt}}
\tblrow{OE\_ORDER\_LINES\_ALL}{24}{446\,M}{\tblbar{figorange}{12.11pt}}
\tblrow{OE\_ORDER\_SOURCES}{12}{40}{\tblbar{figorange}{2.26pt}}
\tblrow{OE\_PRICE\_ADJUSTMENTS}{35}{52.9\,M}{\tblbar{figorange}{10.81pt}}
\tblrow{OE\_TRANSACTION\_TYPES\_ALL}{13}{4}{\tblbar{figorange}{0.98pt}}
\tblrow{OE\_TRANSACTION\_TYPES\_TL}{13}{4}{\tblbar{figorange}{0.98pt}}
\tblmodule{figorange}{Advanced Pricing}{5}{135\,k}{1}
\tblrow{QP\_LIST\_HEADERS\_B}{29}{33.6\,k}{\tblbar{figorange}{6.34pt}}
\tblrow{QP\_LIST\_HEADERS\_TL}{9}{33.6\,k}{\tblbar{figorange}{6.34pt}}
\tblrow{QP\_LIST\_LINES}{33}{33.8\,k}{\tblbar{figorange}{6.34pt}}
\tblrow{QP\_PRICING\_ATTRIBUTES}{23}{33.9\,k}{\tblbar{figorange}{6.34pt}}
\tblrow{QP\_QUALIFIERS}{24}{2}{\tblbar{figorange}{0.67pt}}
\tblmodule{figorange}{Trading Community}{12}{97.5\,M}{1}
\tblrow{HZ\_CODE\_ASSIGNMENTS}{18}{28}{\tblbar{figorange}{2.05pt}}
\tblrow{HZ\_CONTACT\_POINTS}{18}{8.45\,M}{\tblbar{figorange}{9.70pt}}
\tblrow{HZ\_CUSTOMER\_PROFILES}{22}{4.21\,M}{\tblbar{figorange}{9.27pt}}
\tblrow{HZ\_CUST\_ACCOUNTS}{14}{4.21\,M}{\tblbar{figorange}{9.27pt}}
\tblrow{HZ\_CUST\_ACCOUNT\_ROLES}{15}{0}{}
\tblrow{HZ\_CUST\_ACCT\_SITES\_ALL}{17}{18.5\,M}{\tblbar{figorange}{10.17pt}}
\tblrow{HZ\_CUST\_PROFILE\_CLASSES}{21}{1}{\tblbar{figorange}{0.60pt}}
\tblrow{HZ\_CUST\_SITE\_USES\_ALL}{17}{22.7\,M}{\tblbar{figorange}{10.30pt}}
\tblrow{HZ\_LOCATIONS}{34}{16.6\,M}{\tblbar{figorange}{10.11pt}}
\tblrow{HZ\_ORGANIZATION\_PROFILES}{17}{17.9\,k}{\tblbar{figorange}{5.95pt}}
\tblrow{HZ\_PARTIES}{31}{4.31\,M}{\tblbar{figorange}{9.29pt}}
\tblrow{HZ\_PARTY\_SITES}{15}{18.5\,M}{\tblbar{figorange}{10.17pt}}
\tblmodule{figorange}{Contracts Core}{9}{17.2\,M}{1}
\tblrow{OKC\_K\_HEADERS\_ALL\_B}{53}{1.61\,M}{\tblbar{figorange}{8.69pt}}
\tblrow{OKC\_K\_HEADERS\_TL}{11}{1.61\,M}{\tblbar{figorange}{8.69pt}}
\tblrow{OKC\_K\_ITEMS}{21}{2.51\,M}{\tblbar{figorange}{8.96pt}}
\tblrow{OKC\_K\_LINES\_B}{49}{2.51\,M}{\tblbar{figorange}{8.96pt}}
\tblrow{OKC\_K\_LINES\_TL}{11}{2.51\,M}{\tblbar{figorange}{8.96pt}}
\tblrow{OKC\_K\_PARTY\_ROLES\_B}{17}{3.22\,M}{\tblbar{figorange}{9.11pt}}
\tblrow{OKC\_K\_PARTY\_ROLES\_TL}{9}{3.22\,M}{\tblbar{figorange}{9.11pt}}
\tblrow{OKC\_STATUSES\_B}{11}{8}{\tblbar{figorange}{1.34pt}}
\tblrow{OKC\_STATUSES\_TL}{11}{8}{\tblbar{figorange}{1.34pt}}
\tblmodule{figorange}{Service Contracts: subscriptions}{6}{25.7\,M}{1}
\tblrow{OKS\_BILL\_CONT\_LINES}{33}{5.92\,M}{\tblbar{figorange}{9.48pt}}
\tblrow{OKS\_BILL\_TXN\_LINES}{18}{5.92\,M}{\tblbar{figorange}{9.48pt}}
\tblrow{OKS\_K\_LINES\_B}{29}{2.51\,M}{\tblbar{figorange}{8.96pt}}
\tblrow{OKS\_K\_LINES\_TL}{10}{2.51\,M}{\tblbar{figorange}{8.96pt}}
\tblrow{OKS\_LEVEL\_ELEMENTS}{19}{6.31\,M}{\tblbar{figorange}{9.52pt}}
\tblrow{OKS\_STREAM\_LEVELS\_B}{24}{2.51\,M}{\tblbar{figorange}{8.96pt}}
\tblmodule{figorange}{Inventory: items \& catalog}{9}{3.11\,M}{1}
\tblrow{MTL\_CATEGORIES\_B}{10}{70}{\tblbar{figorange}{2.59pt}}
\tblrow{MTL\_CATEGORIES\_TL}{9}{70}{\tblbar{figorange}{2.59pt}}
\tblrow{MTL\_CATEGORY\_SETS\_B}{11}{1}{\tblbar{figorange}{0.60pt}}
\tblrow{MTL\_CATEGORY\_SETS\_TL}{9}{1}{\tblbar{figorange}{0.60pt}}
\tblrow{MTL\_ITEM\_CATEGORIES}{13}{1.03\,M}{\tblbar{figorange}{8.42pt}}
\tblrow{MTL\_PARAMETERS}{12}{17.9\,k}{\tblbar{figorange}{5.95pt}}
\tblrow{MTL\_SYSTEM\_ITEMS\_B}{32}{1.03\,M}{\tblbar{figorange}{8.42pt}}
\tblrow{MTL\_SYSTEM\_ITEMS\_TL}{10}{1.03\,M}{\tblbar{figorange}{8.42pt}}
\tblrow{MTL\_UNITS\_OF\_MEASURE\_TL}{17}{1}{\tblbar{figorange}{0.60pt}}
\tblmodule{figorange}{Human Resources: organizations \& sites}{4}{53.6\,k}{1}
\tblrow{HR\_ALL\_ORGANIZATION\_UNITS}{16}{17.9\,k}{\tblbar{figorange}{5.95pt}}
\tblrow{HR\_LOCATIONS\_ALL}{17}{17.9\,k}{\tblbar{figorange}{5.95pt}}
\tblrow{HR\_OPERATING\_UNITS}{6}{1}{\tblbar{figorange}{0.60pt}}
\tblrow{HR\_ORGANIZATION\_INFORMATION}{14}{17.9\,k}{\tblbar{figorange}{5.95pt}}
\end{tabular}
\end{minipage}\hfill
\begin{minipage}[t]{0.475\linewidth}
\begin{tabular}[t]{@{}T@{}}
\tblfamily{figaqua}{Finance}{79}{4.18\,B}
\tblcolhead
\tblmodule{figaqua}{Receivables}{22}{1.64\,B}{0}
\tblrow{AR\_ADJUSTMENTS\_ALL}{25}{0}{}
\tblrow{AR\_AGING\_BUCKETS}{9}{1}{\tblbar{figaqua}{0.60pt}}
\tblrow{AR\_AGING\_BUCKET\_LINES\_B}{11}{5}{\tblbar{figaqua}{1.09pt}}
\tblrow{AR\_AGING\_BUCKET\_LINES\_TL}{9}{5}{\tblbar{figaqua}{1.09pt}}
\tblrow{AR\_BATCHES\_ALL}{22}{493}{\tblbar{figaqua}{3.77pt}}
\tblrow{AR\_CASH\_RECEIPTS\_ALL}{39}{85.8\,M}{\tblbar{figaqua}{11.11pt}}
\tblrow{AR\_CASH\_RECEIPT\_HISTORY\_ALL}{25}{258\,M}{\tblbar{figaqua}{11.78pt}}
\tblrow{AR\_DISTRIBUTIONS\_ALL}{12}{159\,M}{\tblbar{figaqua}{11.48pt}}
\tblrow{AR\_PAYMENT\_SCHEDULES\_ALL}{29}{174\,M}{\tblbar{figaqua}{11.54pt}}
\tblrow{AR\_RECEIPT\_CLASSES}{12}{1}{\tblbar{figaqua}{0.60pt}}
\tblrow{AR\_RECEIPT\_METHODS}{9}{2}{\tblbar{figaqua}{0.67pt}}
\tblrow{AR\_RECEIVABLES\_TRX\_ALL}{12}{2}{\tblbar{figaqua}{0.67pt}}
\tblrow{AR\_RECEIVABLE\_APPLICATIONS\_ALL}{27}{90.9\,M}{\tblbar{figaqua}{11.14pt}}
\tblrow{RA\_BATCH\_SOURCES\_ALL}{10}{3}{\tblbar{figaqua}{0.84pt}}
\tblrow{RA\_CUSTOMER\_TRX\_ALL}{26}{88.9\,M}{\tblbar{figaqua}{11.13pt}}
\tblrow{RA\_CUSTOMER\_TRX\_LINES\_ALL}{20}{533\,M}{\tblbar{figaqua}{12.22pt}}
\tblrow{RA\_CUST\_TRX\_LINE\_GL\_DIST\_ALL}{20}{246\,M}{\tblbar{figaqua}{11.75pt}}
\tblrow{RA\_CUST\_TRX\_TYPES\_ALL}{20}{4}{\tblbar{figaqua}{0.98pt}}
\tblrow{RA\_INTERFACE\_LINES\_ALL}{18}{0}{}
\tblrow{RA\_TERMS\_B}{12}{3}{\tblbar{figaqua}{0.84pt}}
\tblrow{RA\_TERMS\_LINES}{9}{3}{\tblbar{figaqua}{0.84pt}}
\tblrow{RA\_TERMS\_TL}{10}{3}{\tblbar{figaqua}{0.84pt}}
\tblmodule{figaqua}{Payables}{12}{48.6\,M}{1}
\tblrow{AP\_CHECKS\_ALL}{22}{5.25\,M}{\tblbar{figaqua}{9.41pt}}
\tblrow{AP\_HOLDS\_ALL}{14}{31.2\,k}{\tblbar{figaqua}{6.29pt}}
\tblrow{AP\_HOLD\_CODES}{11}{13}{\tblbar{figaqua}{1.60pt}}
\tblrow{AP\_INVOICES\_ALL}{36}{6.06\,M}{\tblbar{figaqua}{9.50pt}}
\tblrow{AP\_INVOICE\_DISTRIBUTIONS\_ALL}{19}{7.89\,M}{\tblbar{figaqua}{9.66pt}}
\tblrow{AP\_INVOICE\_LINES\_ALL}{17}{17.4\,M}{\tblbar{figaqua}{10.14pt}}
\tblrow{AP\_INVOICE\_PAYMENTS\_ALL}{15}{5.65\,M}{\tblbar{figaqua}{9.45pt}}
\tblrow{AP\_PAYMENT\_SCHEDULES\_ALL}{13}{6.06\,M}{\tblbar{figaqua}{9.50pt}}
\tblrow{AP\_SUPPLIERS}{33}{98.7\,k}{\tblbar{figaqua}{6.99pt}}
\tblrow{AP\_SUPPLIER\_SITES\_ALL}{19}{98.7\,k}{\tblbar{figaqua}{6.99pt}}
\tblrow{AP\_TERMS\_LINES}{9}{3}{\tblbar{figaqua}{0.84pt}}
\tblrow{AP\_TERMS\_TL}{11}{3}{\tblbar{figaqua}{0.84pt}}
\tblmodule{figaqua}{Payments}{6}{11.2\,M}{1}
\tblrow{IBY\_DOCS\_PAYABLE\_ALL}{31}{5.65\,M}{\tblbar{figaqua}{9.45pt}}
\tblrow{IBY\_EXTERNAL\_PAYEES\_ALL}{12}{98.7\,k}{\tblbar{figaqua}{6.99pt}}
\tblrow{IBY\_EXT\_BANK\_ACCOUNTS}{18}{98.1\,k}{\tblbar{figaqua}{6.99pt}}
\tblrow{IBY\_PAYMENTS\_ALL}{34}{5.25\,M}{\tblbar{figaqua}{9.41pt}}
\tblrow{IBY\_PAY\_INSTRUCTIONS\_ALL}{23}{1.07\,k}{\tblbar{figaqua}{4.24pt}}
\tblrow{IBY\_PMT\_INSTR\_USES\_ALL}{14}{101\,k}{\tblbar{figaqua}{7.00pt}}
\tblmodule{figaqua}{Cash Management}{10}{96.1\,M}{1}
\tblrow{CE\_BANK\_ACCOUNTS}{34}{1}{\tblbar{figaqua}{0.60pt}}
\tblrow{CE\_BANK\_ACCT\_USES\_ALL}{21}{1}{\tblbar{figaqua}{0.60pt}}
\tblrow{CE\_GL\_ACCOUNTS\_CCID}{13}{1}{\tblbar{figaqua}{0.60pt}}
\tblrow{CE\_STATEMENT\_HEADERS}{20}{251}{\tblbar{figaqua}{3.36pt}}
\tblrow{CE\_STATEMENT\_HEADERS\_INT}{13}{0}{}
\tblrow{CE\_STATEMENT\_LINES}{19}{5.4\,M}{\tblbar{figaqua}{9.43pt}}
\tblrow{CE\_STATEMENT\_LINES\_INTERFACE}{14}{0}{}
\tblrow{CE\_STATEMENT\_RECONCILS\_ALL}{18}{90.7\,M}{\tblbar{figaqua}{11.14pt}}
\tblrow{CE\_SYSTEM\_PARAMETERS}{16}{1}{\tblbar{figaqua}{0.60pt}}
\tblrow{CE\_TRANSACTION\_CODES}{16}{6}{\tblbar{figaqua}{1.18pt}}
\tblmodule{figaqua}{E-Business Tax}{8}{159\,M}{1}
\tblrow{ZX\_LINES}{21}{79.3\,M}{\tblbar{figaqua}{11.06pt}}
\tblrow{ZX\_LINES\_DET\_FACTORS}{22}{79.3\,M}{\tblbar{figaqua}{11.06pt}}
\tblrow{ZX\_RATES\_B}{18}{1}{\tblbar{figaqua}{0.60pt}}
\tblrow{ZX\_RATES\_TL}{9}{1}{\tblbar{figaqua}{0.60pt}}
\tblrow{ZX\_REGIMES\_B}{13}{1}{\tblbar{figaqua}{0.60pt}}
\tblrow{ZX\_REGIMES\_TL}{9}{1}{\tblbar{figaqua}{0.60pt}}
\tblrow{ZX\_TAXES\_B}{15}{1}{\tblbar{figaqua}{0.60pt}}
\tblrow{ZX\_TAXES\_TL}{9}{1}{\tblbar{figaqua}{0.60pt}}
\end{tabular}
\end{minipage}
\end{table}

\begin{table}[p]
\tblfont\setlength{\tabcolsep}{2.2pt}\renewcommand{\arraystretch}{1}
{\small\textit{Table~\ref{tab:tables}, continued.}}\par\medskip
\centering
\begin{minipage}[t]{0.475\linewidth}
\begin{tabular}[t]{@{}T@{}}
\tblfamily{figaqua}{Finance \textmd{\textit{(cont.)}}}{79}{4.18\,B}
\tblcolhead
\tblmodule{figaqua}{Subledger Accounting}{7}{1.33\,B}{0}
\tblrow{XLA\_AE\_HEADERS}{22}{167\,M}{\tblbar{figaqua}{11.51pt}}
\tblrow{XLA\_AE\_LINES}{25}{413\,M}{\tblbar{figaqua}{12.06pt}}
\tblrow{XLA\_DISTRIBUTION\_LINKS}{7}{413\,M}{\tblbar{figaqua}{12.06pt}}
\tblrow{XLA\_EVENTS}{18}{167\,M}{\tblbar{figaqua}{11.51pt}}
\tblrow{XLA\_EVENT\_TYPES\_B}{12}{5}{\tblbar{figaqua}{1.09pt}}
\tblrow{XLA\_EVENT\_TYPES\_TL}{13}{5}{\tblbar{figaqua}{1.09pt}}
\tblrow{XLA\_TRANSACTION\_ENTITIES}{12}{166\,M}{\tblbar{figaqua}{11.51pt}}
\tblmodule{figaqua}{General Ledger}{14}{901\,M}{1}
\tblrow{GL\_ACCOUNT\_HIERARCHIES}{7}{49.3\,k}{\tblbar{figaqua}{6.57pt}}
\tblrow{GL\_BALANCES}{22}{204\,k}{\tblbar{figaqua}{7.43pt}}
\tblrow{GL\_CODE\_COMBINATIONS}{19}{10.7\,k}{\tblbar{figaqua}{5.64pt}}
\tblrow{GL\_IMPORT\_REFERENCES}{17}{899\,M}{\tblbar{figaqua}{12.53pt}}
\tblrow{GL\_INTERFACE}{17}{0}{}
\tblrow{GL\_JE\_BATCHES}{15}{3.07\,k}{\tblbar{figaqua}{4.88pt}}
\tblrow{GL\_JE\_CATEGORIES\_TL}{10}{9}{\tblbar{figaqua}{1.40pt}}
\tblrow{GL\_JE\_HEADERS}{19}{4.38\,k}{\tblbar{figaqua}{5.10pt}}
\tblrow{GL\_JE\_LINES}{17}{1.82\,M}{\tblbar{figaqua}{8.76pt}}
\tblrow{GL\_JE\_SOURCES\_TL}{15}{9}{\tblbar{figaqua}{1.40pt}}
\tblrow{GL\_LEDGERS}{45}{1}{\tblbar{figaqua}{0.60pt}}
\tblrow{GL\_PERIODS}{17}{26}{\tblbar{figaqua}{2.00pt}}
\tblrow{GL\_PERIOD\_STATUSES}{20}{104}{\tblbar{figaqua}{2.83pt}}
\tblrow{GL\_SUMMARY\_TEMPLATES}{21}{3}{\tblbar{figaqua}{0.84pt}}
\tblfamily{figblue}{Custom extensions}{76}{2.59\,B}
\tblmodule{figblue}{Order lifecycle \& kitchens}{7}{779\,M}{0}
\tblrow{XX\_ORDER\_STATUS\_HIST}{13}{483\,M}{\tblbar{figblue}{12.16pt}}
\tblrow{XX\_ORDER\_PROMISES}{3}{71.6\,M}{\tblbar{figblue}{11.00pt}}
\tblrow{XX\_ORDER\_PREP\_QUOTES}{16}{81\,M}{\tblbar{figblue}{11.07pt}}
\tblrow{XX\_ORDER\_REJECTIONS}{18}{438\,k}{\tblbar{figblue}{7.90pt}}
\tblrow{XX\_PREP\_ESTIMATE\_LOG}{18}{35.8\,M}{\tblbar{figblue}{10.58pt}}
\tblrow{XX\_MERCHANT\_READY\_REPORTS}{20}{79.2\,M}{\tblbar{figblue}{11.06pt}}
\tblrow{XX\_MERCHANT\_HANDOFF\_CONFIRMATIONS}{6}{27.9\,M}{\tblbar{figblue}{10.42pt}}
\tblmodule{figblue}{Dispatch \& delivery}{10}{431\,M}{1}
\tblrow{XX\_DISPATCH\_ASSIGNMENTS}{18}{125\,M}{\tblbar{figblue}{11.34pt}}
\tblrow{XX\_DELIVERY\_ASSIGNMENTS}{11}{72.9\,M}{\tblbar{figblue}{11.01pt}}
\tblrow{XX\_DELIVERY\_LEGS}{26}{69.6\,M}{\tblbar{figblue}{10.98pt}}
\tblrow{XX\_DELIVERY\_ATTEMPTS}{17}{1.96\,M}{\tblbar{figblue}{8.81pt}}
\tblrow{XX\_COURIER\_WAITS}{22}{21.7\,M}{\tblbar{figblue}{10.27pt}}
\tblrow{XX\_DRIVER\_LOCATION\_OBSERVATIONS}{8}{134\,M}{\tblbar{figblue}{11.38pt}}
\tblrow{XX\_HOTSPOT\_SNAPSHOTS}{7}{109\,k}{\tblbar{figblue}{7.05pt}}
\tblrow{XX\_SUPPLY\_SNAPSHOTS}{8}{1.98\,M}{\tblbar{figblue}{8.81pt}}
\tblrow{XX\_SUPPLY\_FORECASTS}{22}{569\,k}{\tblbar{figblue}{8.06pt}}
\tblrow{XX\_MARKETPLACE\_ESTIMATES}{22}{2.64\,M}{\tblbar{figblue}{8.99pt}}
\tblmodule{figblue}{Couriers}{8}{12.7\,M}{1}
\tblrow{XX\_DRIVER\_PROFILES}{18}{80.9\,k}{\tblbar{figblue}{6.87pt}}
\tblrow{XX\_COURIER\_APPLICATIONS}{8}{39.9\,k}{\tblbar{figblue}{6.44pt}}
\tblrow{XX\_COURIER\_IDENTITY\_CHECKS}{11}{154\,k}{\tblbar{figblue}{7.26pt}}
\tblrow{XX\_COURIER\_ACCOUNT\_EVENTS}{8}{9.77\,k}{\tblbar{figblue}{5.59pt}}
\tblrow{XX\_COURIER\_DEVICES}{6}{115\,k}{\tblbar{figblue}{7.09pt}}
\tblrow{XX\_COURIER\_APP\_SESSIONS}{9}{6.06\,M}{\tblbar{figblue}{9.50pt}}
\tblrow{XX\_DRIVER\_SHIFTS}{15}{6.09\,M}{\tblbar{figblue}{9.50pt}}
\tblrow{XX\_COURIER\_VIOLATIONS}{9}{113\,k}{\tblbar{figblue}{7.07pt}}
\tblmodule{figblue}{Courier pay \& incentives}{9}{141\,M}{1}
\tblrow{XX\_COURIER\_EARNINGS}{15}{69.6\,M}{\tblbar{figblue}{10.98pt}}
\tblrow{XX\_TIP\_ADJUSTMENTS}{14}{164\,k}{\tblbar{figblue}{7.30pt}}
\tblrow{XX\_PAYOUT\_PERIODS}{16}{499}{\tblbar{figblue}{3.78pt}}
\tblrow{XX\_PAY\_PERIOD\_INCENTIVES}{32}{53}{\tblbar{figblue}{2.43pt}}
\tblrow{XX\_INCENTIVE\_COMMITMENTS}{26}{51.1\,M}{\tblbar{figblue}{10.79pt}}
\tblrow{XX\_DRIVER\_QUESTS}{16}{138\,k}{\tblbar{figblue}{7.19pt}}
\tblrow{XX\_QUEST\_WINDOWS}{14}{138\,k}{\tblbar{figblue}{7.19pt}}
\tblrow{XX\_QUEST\_INVITATIONS}{14}{15.1\,M}{\tblbar{figblue}{10.05pt}}
\tblrow{XX\_QUEST\_PROGRESS}{17}{4.99\,M}{\tblbar{figblue}{9.38pt}}
\tblmodule{figblue}{Merchants}{8}{15.2\,M}{1}
\tblrow{XX\_MERCHANT\_HOURS}{15}{125\,k}{\tblbar{figblue}{7.14pt}}
\tblrow{XX\_MERCHANT\_HOLIDAY\_HOURS}{14}{87.5\,k}{\tblbar{figblue}{6.92pt}}
\tblrow{XX\_MERCHANT\_CLOSURES}{17}{364\,k}{\tblbar{figblue}{7.79pt}}
\tblrow{XX\_MERCHANT\_INTEGRATIONS}{23}{17.9\,k}{\tblbar{figblue}{5.95pt}}
\tblrow{XX\_MERCHANT\_OPS\_ACTIONS}{26}{3.12\,k}{\tblbar{figblue}{4.89pt}}
\tblrow{XX\_MERCHANT\_RATING\_SNAPSHOTS}{14}{5.06\,M}{\tblbar{figblue}{9.39pt}}
\tblrow{XX\_MERCHANT\_REGULATORY\_EVENTS}{15}{12.3\,k}{\tblbar{figblue}{5.73pt}}
\tblrow{XX\_REVIEWS}{16}{9.52\,M}{\tblbar{figblue}{9.77pt}}
\end{tabular}
\end{minipage}\hfill
\begin{minipage}[t]{0.475\linewidth}
\begin{tabular}[t]{@{}T@{}}
\tblfamily{figblue}{Custom extensions \textmd{\textit{(cont.)}}}{76}{2.59\,B}
\tblcolhead
\tblmodule{figblue}{Promotions, membership \& referrals}{14}{275\,M}{0}
\tblrow{XX\_PROMO\_CODES}{20}{33.8\,k}{\tblbar{figblue}{6.34pt}}
\tblrow{XX\_PROMO\_ATTEMPTS}{20}{33.1\,M}{\tblbar{figblue}{10.53pt}}
\tblrow{XX\_PROMOTION\_VARIANTS}{14}{215}{\tblbar{figblue}{3.27pt}}
\tblrow{XX\_PROMOTION\_DECISIONS}{22}{13.6\,k}{\tblbar{figblue}{5.79pt}}
\tblrow{XX\_PROMOTION\_CONTACTS}{12}{17.1\,M}{\tblbar{figblue}{10.13pt}}
\tblrow{XX\_PROMOTION\_SUBMISSIONS}{8}{40.1\,M}{\tblbar{figblue}{10.64pt}}
\tblrow{XX\_PROMOTION\_ALLOCATIONS}{18}{19\,M}{\tblbar{figblue}{10.19pt}}
\tblrow{XX\_PROMOTION\_CHECKOUTS}{11}{99.2\,M}{\tblbar{figblue}{11.19pt}}
\tblrow{XX\_CAMPAIGN\_AUTHORITY\_EVENTS}{15}{85.1\,k}{\tblbar{figblue}{6.90pt}}
\tblrow{XX\_CAMPAIGN\_BUDGET\_CREDITS}{20}{34.4\,k}{\tblbar{figblue}{6.35pt}}
\tblrow{XX\_EXPERIMENT\_ASSIGNMENTS}{24}{8.1\,k}{\tblbar{figblue}{5.47pt}}
\tblrow{XX\_MEMBERSHIP\_EVENTS}{19}{21.2\,M}{\tblbar{figblue}{10.26pt}}
\tblrow{XX\_MEMBERSHIP\_ORDER\_BENEFITS}{19}{45\,M}{\tblbar{figblue}{10.71pt}}
\tblrow{XX\_REFERRALS}{15}{507\,k}{\tblbar{figblue}{7.99pt}}
\tblmodule{figblue}{Accounts, devices \& payment risk}{12}{443\,M}{1}
\tblrow{XX\_CUSTOMER\_ACCOUNT\_EVENTS}{20}{817\,k}{\tblbar{figblue}{8.28pt}}
\tblrow{XX\_CUSTOMER\_SESSIONS}{16}{284\,M}{\tblbar{figblue}{11.83pt}}
\tblrow{XX\_LOGIN\_ATTEMPTS}{15}{33.2\,M}{\tblbar{figblue}{10.53pt}}
\tblrow{XX\_DEVICES}{13}{9.41\,M}{\tblbar{figblue}{9.76pt}}
\tblrow{XX\_CUSTOMER\_DEVICES}{13}{9.79\,M}{\tblbar{figblue}{9.79pt}}
\tblrow{XX\_IP\_ADDRESSES}{17}{2.66\,M}{\tblbar{figblue}{9.00pt}}
\tblrow{XX\_PAYMENT\_INSTRUMENTS}{16}{7.36\,M}{\tblbar{figblue}{9.61pt}}
\tblrow{XX\_CARD\_FINGERPRINTS}{12}{7.38\,M}{\tblbar{figblue}{9.62pt}}
\tblrow{XX\_CARD\_VERIFICATIONS}{21}{2.68\,M}{\tblbar{figblue}{9.00pt}}
\tblrow{XX\_PAYMENT\_AUTHS}{16}{86.1\,M}{\tblbar{figblue}{11.11pt}}
\tblrow{XX\_FRAUD\_WARNINGS}{16}{24.4\,k}{\tblbar{figblue}{6.14pt}}
\tblrow{XX\_ENFORCEMENT\_ACTIONS}{17}{11\,k}{\tblbar{figblue}{5.66pt}}
\tblmodule{figblue}{Support, refunds \& disputes}{3}{9.17\,M}{1}
\tblrow{XX\_SUPPORT\_CASES}{30}{7.24\,M}{\tblbar{figblue}{9.60pt}}
\tblrow{XX\_REFUNDS}{20}{1.88\,M}{\tblbar{figblue}{8.78pt}}
\tblrow{XX\_DISPUTES}{17}{59\,k}{\tblbar{figblue}{6.68pt}}
\tblmodule{figblue}{Integration \& audit}{5}{486\,M}{1}
\tblrow{XX\_INGESTION\_RUNS}{19}{16.9\,k}{\tblbar{figblue}{5.92pt}}
\tblrow{XX\_INTERFACE\_ERRORS}{20}{2.23\,k}{\tblbar{figblue}{4.69pt}}
\tblrow{XX\_INTEGRATION\_INCIDENTS}{15}{24\,k}{\tblbar{figblue}{6.13pt}}
\tblrow{XX\_GL\_INTERFACE\_HIST}{9}{486\,M}{\tblbar{figblue}{12.16pt}}
\tblrow{XX\_AUDIT\_TRAIL}{21}{282\,k}{\tblbar{figblue}{7.63pt}}
\tblfamily{figgrey}{System}{29}{17.4\,k}
\tblmodule{figgrey}{Application Object Library}{29}{17.4\,k}{0}
\tblrow{FND\_APPLICATION}{9}{12}{\tblbar{figgrey}{1.56pt}}
\tblrow{FND\_APPLICATION\_TL}{9}{12}{\tblbar{figgrey}{1.56pt}}
\tblrow{FND\_CONCURRENT\_PROGRAMS}{10}{7}{\tblbar{figgrey}{1.26pt}}
\tblrow{FND\_CONCURRENT\_PROGRAMS\_TL}{11}{7}{\tblbar{figgrey}{1.26pt}}
\tblrow{FND\_CONCURRENT\_REQUESTS}{13}{15.8\,k}{\tblbar{figgrey}{5.88pt}}
\tblrow{FND\_CURRENCIES}{11}{68}{\tblbar{figgrey}{2.57pt}}
\tblrow{FND\_CURRENCIES\_TL}{9}{68}{\tblbar{figgrey}{2.57pt}}
\tblrow{FND\_FLEX\_HIERARCHIES}{8}{2}{\tblbar{figgrey}{0.67pt}}
\tblrow{FND\_FLEX\_HIERARCHIES\_TL}{11}{2}{\tblbar{figgrey}{0.67pt}}
\tblrow{FND\_FLEX\_VALIDATION\_QUALIFIERS}{6}{5}{\tblbar{figgrey}{1.09pt}}
\tblrow{FND\_FLEX\_VALUES}{18}{351}{\tblbar{figgrey}{3.57pt}}
\tblrow{FND\_FLEX\_VALUES\_TL}{10}{351}{\tblbar{figgrey}{3.57pt}}
\tblrow{FND\_FLEX\_VALUE\_HIERARCHIES}{11}{33}{\tblbar{figgrey}{2.14pt}}
\tblrow{FND\_FLEX\_VALUE\_NORM\_HIERARCHY}{12}{26}{\tblbar{figgrey}{2.00pt}}
\tblrow{FND\_FLEX\_VALUE\_SETS}{11}{7}{\tblbar{figgrey}{1.26pt}}
\tblrow{FND\_ID\_FLEXS}{14}{1}{\tblbar{figgrey}{0.60pt}}
\tblrow{FND\_ID\_FLEX\_SEGMENTS}{17}{7}{\tblbar{figgrey}{1.26pt}}
\tblrow{FND\_ID\_FLEX\_SEGMENTS\_TL}{13}{7}{\tblbar{figgrey}{1.26pt}}
\tblrow{FND\_ID\_FLEX\_STRUCTURES}{14}{1}{\tblbar{figgrey}{0.60pt}}
\tblrow{FND\_ID\_FLEX\_STRUCTURES\_TL}{12}{1}{\tblbar{figgrey}{0.60pt}}
\tblrow{FND\_LANGUAGES}{10}{29}{\tblbar{figgrey}{2.07pt}}
\tblrow{FND\_LOOKUP\_TYPES}{10}{74}{\tblbar{figgrey}{2.63pt}}
\tblrow{FND\_LOOKUP\_TYPES\_TL}{12}{74}{\tblbar{figgrey}{2.63pt}}
\tblrow{FND\_LOOKUP\_VALUES}{13}{292}{\tblbar{figgrey}{3.45pt}}
\tblrow{FND\_RESPONSIBILITY}{12}{8}{\tblbar{figgrey}{1.34pt}}
\tblrow{FND\_RESPONSIBILITY\_TL}{10}{8}{\tblbar{figgrey}{1.34pt}}
\tblrow{FND\_TERRITORIES}{7}{78}{\tblbar{figgrey}{2.66pt}}
\tblrow{FND\_TERRITORIES\_TL}{9}{78}{\tblbar{figgrey}{2.66pt}}
\tblrow{FND\_USER}{10}{43}{\tblbar{figgrey}{2.30pt}}
\end{tabular}
\end{minipage}
\end{table}

%% file: sec/appendix/D_reward_hacking.tex
\section{Reward Hacking}
\label{app:reward-hacking}

In an early round of runs, the Python sandbox did not isolate the agent from the evaluation host. In three runs of fraud-04, Gemini 3.8 Flash changed its root directory, located the grader for the task, and filed the answer it read there, scoring 0.99~\citep{skalse2022defining}. We searched the traces of every run in that round for the same behavior and found it in no other model. The round was discarded. Every run reported in this paper executes Python in a network-less gVisor pod under Inspect's Kubernetes sandbox provider on GKE, and each run's BigQuery service account can read only the dataset scoped to its task. The relevant agent traces are included in the GitHub release.

%% file: sec/appendix/E_system_prompt.tex
\FloatBarrier
\section{System prompt}
\label{app:prompt}

\begin{lstlisting}[style=prompt]
You are an analyst for a food delivery platform operating in NYC. You have read-only access to a transformed export of the company's Oracle E-Business Suite warehouse.
You should explore the warehouse, investigate, and then perform appropriate actions based on your findings.

(*@\textrm{\textbf{Warehouse}}@*)

One Oracle EBS instance, order-to-cash, covering the calendar year 2024, following Oracle house conventions.
Timestamps are naive America/New_York: compare "as of 2024-07-31 23:59:59 ET" directly against stored values, without converting.

(*@\textrm{\textbf{Mission Control}}@*)

File every finding from Python (`run_python`) through the Mission Control console. A number stated only in prose is not filed.
Example:
```python
from mission_control import MissionControl, Reason
mission_control = MissionControl()

# a list can mix bare ids with per-item mappings, with the call's reason= overriding bare ones
mission_control.ban_customers(
    [{"id": 101, "reason": Reason.PROMO_FARMING}, 102],
    reason=Reason.CARD_TESTING,
)

# finish with a summary
mission_control.summary()
```

(*@\textrm{\textbf{\textit{Endpoints}}}@*)

`help(MissionControl.<endpoint>)` shows its docstring and arguments.
Pass the complete set in one call, endpoints are plural.
A rejected call files nothing and says which item is wrong.

(*@\textrm{\textit{Enforcement}}@*)
- `ban_customers`, `ban_couriers`, `ban_merchants`: Close customer accounts, deactivate couriers, remove merchants.
- `hold_payouts`: Freeze a supplier's payouts pending review.
- `discontinue_promo_codes`: Stop promo codes from being redeemed again.
- `flag_rings`: Report sets of actors believed to be operating together.
- `end_campaigns`: Stop promotional campaigns.

(*@\textrm{\textit{Payments and pay}}@*)
- `remediate_payments`: Refund, recapture, void or write off payments on orders.
- `report_pay_periods`: State couriers' statutory minimum-pay position per pay period.
- `issue_pay_adjustments`: Issue pay corrections to couriers.
- `elect_methods`: Record which minimum-pay method was elected per period, with both costs.

(*@\textrm{\textit{Accounting}}@*)
- `report_balances`: State account balances as of an instant.
- `post_journal_entries`: Post balanced journal entries.
- `file_adjustments`: Record reconciling differences against an account.
- `report_rollforwards`: File an account roll-forward, one item per period.
- `report_agings`: Break balances into buckets, by age or any other dimension.
- `flag_orders`: Flag orders for an accounting or control defect.
- `file_schedules`: File the tables a close produces.

(*@\textrm{\textit{Analysis and planning}}@*)
- `report_campaign_performances`: Record measured campaign economics.
- `report_metrics`: Any other figure you were asked for.
- `file_forecasts`: Forecast figures that are not yet knowable, as a point and an interval.
- `file_policies`: Propose operating policies as code.

(*@\textrm{\textit{Dashboards}}@*)
- `data_source_contracts`: The data sources this workspace's dashboard tiles are waiting for.
- `publish_data_sources`: Publish dashboard data sources. This endpoint validates your data source against the tile's contract.

(*@\textrm{\textit{Session}}@*)
- `note`: Attach free-text rationale or caveats that are not a decision.
- `status`: What has been filed so far.
- `summary`: Finish with this.

Money is compared to the cent. Round only at the end. Never modify the warehouse.
\end{lstlisting}

%% file: sec/appendix/F_agent_tools.tex
\section{Agent tools}
\label{app:tools}

The reference agent is given four tools, served by two MCP servers~\citep{anthropic2024mcp}. A read-only \emph{warehouse} server runs over the task's database engine, and a \emph{Python} server holds a persistent interpreter in the agent's sandbox. Filings to Mission Control go through \texttt{mission\_control.py}, which is importable from \texttt{run\_python} (Appendix~\ref{app:prompt}).

\newcommand{\ph}[1]{\textrm{\textit{$\langle$#1$\rangle$}}}

\toolcard{list\_tables}{Warehouse}
  {list\_tables() -> str}
  {List every table in the warehouse.}
  {list\_tables()}

\toolcard{describe\_table}{Warehouse}
  {describe\_table(table: str) -> str}
  {Show a table's columns in order, with types and nullability.}
  {describe\_table(table="OE\_ORDER\_HEADERS\_ALL")}

\toolcard{run\_sql}{Warehouse}
  {run\_sql(sql: str) -> str}
  {Run one read-only SQL query (\ph{SQL dialect}). Unqualified table names resolve to \ph{dataset}. Returns the first 50 rows; the complete result (up to 1,000,000 rows) is saved as a Parquet file whose path is printed last for loading in \texttt{run\_python}.}
  {run\_sql(sql="SELECT ORDER\_NUMBER, ORDERED\_DATE FROM OE\_ORDER\_HEADERS\_ALL WHERE FLOW\_STATUS\_CODE = \textquotesingle CANCELLED\textquotesingle")}

\toolcard{run\_python}{Python}
  {run\_python(code: str) -> str}
  {Run Python in a persistent interpreter: variables, imports, and loaded data carry over between calls, as in a notebook. Returns everything printed, the value of a final bare expression, and any traceback.\par\smallskip
   The working directory holds \texttt{results/}, where every \texttt{run\_sql} result is saved as Parquet (\texttt{pandas.read\_parquet("results/sql\_0003.parquet")}), and \texttt{mission\_control.py}, so \texttt{from mission\_control import MissionControl, Reason, get\_sandbox\_id} works here. pandas, numpy, polars, scipy, statsmodels, scikit-learn, matplotlib, seaborn and Google OR-Tools (\texttt{ortools}: linear, integer and constraint programming, routing) are installed.}
  {run\_python(code="import pandas as pd\newline
   df = pd.read\_parquet(\textquotedbl results/sql\_0001.parquet\textquotedbl)\newline
   df.groupby(\textquotedbl ORDERED\_DATE\textquotedbl).size()")}

\paragraph{Limits.} Each sandbox runs under gVisor~\citep{young2019true} with 4 GiB of memory (12 GiB for the few runs repeated after the sandbox ran out of memory), one CPU, and 10~GiB of ephemeral disk. Each run is limited to 500 model turns and 240 minutes. A \texttt{run\_python} cell times out after 900 seconds and returns at most 30,000 characters of output, and a \texttt{run\_sql} query may scan at most 20~GiB, with no limit on the number of queries, and saves at most 1,000,000 rows. A run that reaches a limit, exhausts the model's context window, or crashes its sandbox is graded on whatever it submitted before stopping. Of the 9,869 graded runs, 56 (0.6\%) reached the turn limit, 43 of them from Muse Spark 1.3 and 13 from Gemini 3.8 Flash, and 65 (0.7\%) reached the context window, all from Qwen 3.8 Max.

%% file: sec/appendix/G_grading.tex
\FloatBarrier
\section{Task definition and grading}
\label{app:grading}

\paragraph{Task definition.} A task consists of a prompt, a scope, a list of expectations, and an answer key. The prompt is the only text the agent receives besides the system prompt (Appendix~\ref{app:prompt}). The scope sets the last month of 2024 in the warehouse the agent queries. Each expectation names one filing the grader requires, with its action (for instance \texttt{ban\_couriers} or \texttt{file\_forecasts}), the keys it is filed against, such as vendor IDs or months, and its grading mode. The answer key is computed once, before any run, and frozen. It comes either from truth queries, which are SQL over the latent tables, or from a local builder, which reads the simulator's labels directly, for instance which couriers were made to steal orders. Listing~\ref{lst:task} gives both definitions with their defaults.

\begin{lstlisting}[style=cardpy,numbers=none,xleftmargin=6pt,framexleftmargin=6pt,float=!ht,
  deletekeywords={id}, emph={Task,Expectation,DataSource,PolicySpec},
  caption={The \texttt{Task} and \texttt{Expectation} definitions of the harness, with their defaults.},
  label={lst:task}]
Task(id, title, suite, prompt,
     as_of="",             # maturity cutoff for the answer key
     month="",             # last month of 2024 the agent's warehouse view shows
     key_month=None,       # warehouse the answer key is computed from, if not `month`
     expects=(),           # filing kinds the task requires, checked by lint only
     expectations=(),      # the scoreable claims
     truth_queries={},     # section name -> SQL, run against BigQuery and frozen
     truth_labels={},      # sections read from the world's label sidecar instead
     pins=(),              # property-pinned entities; the prompt becomes a template
     data_sources=(),      # DataSource contracts, for `data_source` mode
     derive_truth=None,    # callable: truth -> more truth sections
     lab="",               # policy_arena bundle, for `policy*` modes
     notes="")             # provenance; where every measured figure is recorded

Expectation(metric, kind, mode,
     truth_key="",         # which answer-key section
     id_key="",            # the payload field carrying the row's identity
     value_key="amount",   # the payload field carrying the figure
     tolerance=0.01, rel_tolerance=0.0,
     penalize_extras=False,
     schedule=None,        # table mode: which filed schedule to grade
     weight=1.0, optional=False, note="",
     # forecast:    horizon, history_through, reference, season, ceiling
     # cost_set:    cost_key, margin_key, admin_cost, verdict_key,
     #              margin_retention, neutral_review_share
     # data_source: source;  policy*: policy=PolicySpec(...)
     )
\end{lstlisting}

\paragraph{Grading modes.} Table~\ref{tab:graders} lists the nine grading modes and the actions agents file.

\begin{table}[!ht]
\caption{Grading modes (left) and filed actions (right) over the \ntasks tasks. A task counts once for each mode or action it uses, so the columns sum to more than the number of tasks.}
\label{tab:graders}
\centering
\footnotesize
\begin{minipage}[t]{0.57\linewidth}\vspace{0pt}
\begin{tabularx}{\linewidth}{@{}l>{\raggedright\arraybackslash}Xr@{}}
\toprule
Mode & Scored by & Tasks \\
\midrule
\texttt{forecast} & Weighted interval score, on a scale set by a reference forecast & 72 \\
\texttt{cost\_set} & Cost saved by bans or holds (Equation~\ref{eq:cost}) & 43 \\
\texttt{data\_source} & Structural contract, then every value of the tile & 36 \\
\texttt{presence} & Whether a filing of that kind was made, such as a written note & 34 \\
\texttt{id\_set} & Overlap of a set of IDs with the key (F1) & 28 \\
\texttt{keyed\_value} & Values per key, each within a tolerance & 22 \\
\texttt{table} & A filed table, rows matched on its key columns and each value within a tolerance & 14 \\
\texttt{allocation} & Share of attainable savings realized in the simulator & 9 \\
\texttt{binary} & All or nothing: exactly the right set of IDs & 1 \\
\bottomrule
\end{tabularx}
\end{minipage}\hfill
\begin{minipage}[t]{0.40\linewidth}\vspace{0pt}
\begin{tabular}{@{}lr@{}}
\toprule
Action & Tasks \\
\midrule
\texttt{file\_forecasts} & 72 \\
\texttt{publish\_data\_sources} & 36 \\
\texttt{note} & 34 \\
\texttt{ban\_couriers} & 26 \\
\texttt{report\_metrics} & 24 \\
\texttt{file\_schedules} & 23 \\
\texttt{ban\_customers} & 22 \\
\texttt{hold\_payouts} & 12 \\
\texttt{ban\_merchants} & 10 \\
\texttt{remediate\_payments} & 3 \\
\texttt{report\_pay\_periods} & 1 \\
\texttt{discontinue\_promo\_codes} & 1 \\
\bottomrule
\end{tabular}
\end{minipage}
\end{table}

\paragraph{Task score.} Each expectation scores between 0 and 100, a forecast between $-200$ and 100, and the task score is their weighted mean. Mixing binary, graded, and forecast scales in one mean is a choice, so Table~\ref{tab:results} also reports the solved rate, which does not depend on how the scales are mixed, and per-domain means, and it floors each model's forecast tasks at 0 as a block so that a few very wrong forecasts cannot outweigh the rest. Of the \ntasks tasks, 159 have one scored expectation, 37 have two, and 14 have three. An expectation marked optional is graded and reported but left out of the mean. If a run files nothing while the key expects at least one action, every expectation scores 0, and a forecast expectation $-200$. A filing that carries many items, such as one \texttt{ban\_customers} call with 300 accounts, is split into its items before grading, so it scores the same as 300 separate filings. IDs and keys are compared as normalized text: case and surrounding whitespace do not matter, and \texttt{6057444.0} is \texttt{6057444}. A key filed inside a longer string, such as \texttt{payout period P17}, also counts, but only on a token boundary, so \texttt{P17} does not match \texttt{P170}.

\paragraph{Forecasts.} A forecast files a point and an 80\% interval for each series and is scored by its weighted interval score (WIS)~\citep{bracher2021evaluating}, with the median and one interval, on a scale set by a reference forecast. For 76 of the 98 forecast expectations the reference is the no-change forecast from the months the agent can see: its point $r$ is the last visible value and its interval is $r \pm z s$ with $s = \hat\sigma\sqrt{h}$, where $\hat\sigma$ is the root mean square of the one-step changes and $h$ the number of periods ahead~\citep{hyndman2014forecasting}. The other 22 cross a regime change, such as the minimum-pay rule, where no change is not a serious benchmark, and their reference is frozen into the answer key as a point $r$ and a standard deviation $s$ at the horizon. Most are a figure the prompt attributes to the business and asks the agent to check, computed by the method the prompt states from the data the agent can see. Table~\ref{tab:fixed-refs} lists all 22 and how each was set, including seven whose point was set with the outcome in view.

Every filed quantile $q$ and the outcome $y$ are mapped to $g(v) = \operatorname{asinh}\big((v - r)/s\big)$, which is linear within about one reference standard deviation of $r$ and grows like a logarithm beyond it, and a series scores
\[
  100\left(1 - \frac{\text{WIS}\big(g(q), g(y)\big)}{W}\right),
\]
where $W = 0.623$ is the median of the references' own WIS on that scale over the 98 series of the benchmark. A perfect forecast scores 100, and a forecast as far from the outcome as the median reference scores 0. Since $r$ and $s$ are fixed before the outcome and $W$ is one constant for the whole benchmark, no series is weighted by how well its own reference happened to do, and since an increasing transformation preserves quantiles, the score is proper: a forecaster maximizes its expected score by filing its own median and 80\% interval~\citep{gneiting2011quantiles, bosse2023transformed}. An expectation averages its series and is floored at $-200$, so one failed forecast costs at most what two perfect ones earn, and a series the agent does not file scores $-200$. The floor is the one improper element; Appendix~\ref{app:findings} measures how little it bends the incentive. In every mean we report, a setting's forecast tasks are floored at 0 as a block, so a model no better than the median reference across its forecasts scores 0 on them rather than below. At the settings of Table~\ref{tab:results} this floor binds for eight of the fourteen models, all but Claude Opus 5.5, GPT-6 Astra, Claude Sonnet 5.5, GPT-6.1 Sol, GPT-6 Sol, and Kimi K3, so the forecasting column separates the strongest models and not the rest.

\begin{table}[!ht]
\caption{The 22 fixed forecast references. The point is the figure the grader centers the scale on and $s$ its standard deviation at the horizon. Rule: computed from the data the agent can see, by the method stated. Hand: a standard deviation set by the authors. The last seven points were set with the realized value in view; these tasks stay in the benchmark, and Appendix~\ref{app:findings} reports the results without them.}
\label{tab:fixed-refs}
\centering
\scriptsize
\begin{tabularx}{\linewidth}{@{}>{\raggedright\arraybackslash}p{0.27\linewidth}>{\raggedright\arraybackslash}X>{\raggedright\arraybackslash}p{0.2\linewidth}@{}}
\toprule
Task (expectations) & Point & $s$ \\
\midrule
\texttt{fc-08} March, hinted and not (2) & The minimum-pay rule applied to the March payout periods, scaled to May's & Rule: spread of the March weeks \\
\texttt{fc-13} April, July (2) & The aggregate floor gap from the plans in the prompt and last month's pay per delivery & Rule: last month's change in pay per delivery \\
\texttt{fc-10h} June hours (1) & The dispatcher's plan: five periods at the mean of the last four & Hand \\
\texttt{fc-11h} June pay (1) & FP\&A's plan: five periods at the mean of the last eight & Hand \\
\texttt{fc-10h}, \texttt{fc-11h} Q3 (2) & The latest week carried through thirteen periods, less 1.3\% a week for hours; chosen after looser plans were beaten & Hand \\
\texttt{fc-14} back pay (1) & Engineering's estimate: dropped trip hours priced at \$19.56 in the weeks the top-up ran & Hand \\
\texttt{fc-14} hard, later weeks (1) & Engineering's figure: nothing owed & Hand \\
\texttt{fc-20} no hint, hard (4) & The quest desk's claims: surge saved as accepted quests at the month's surge premium per hour (split by daypart in proportion to quests run), and cancellations cut by a fifth & Hand \\
\texttt{fc-20p} cancellations (1) & The quest desk's claim & Hand \\
\midrule
\texttt{cb-01p}, \texttt{mr-01p}, \texttt{sw-12p} (4) & Realized value $\times$ 1.024 to 1.026 & About 2.5\% of it \\
\texttt{fc-20p} surge (1) & Realized holdout contrast $\times$ 1.03 & Its standard error \\
\texttt{fc-14} hard, first two weeks (1) & The simulator's label for the aggregate lapse & Hand \\
\texttt{cb-02} (1) & October's open disputes grown by September's losses through December & Hand \\
\bottomrule
\end{tabularx}
\end{table}

\paragraph{Cost savings for bans.} Money saved is calculated against the baselines of banning nobody and banning everybody, to discourage both false positives and false negatives. Every account in the task's universe is a target $T$, an innocent $I$, or a neutral account that is bad but not what the task asks for. Every ban filed carries a review cost $a$, so even correct bans are not free. A missed target $u$ costs its losses $\ell_u$, a wrongly banned innocent costs $a$ plus the margin $m_u$ the platform earns from it, a banned neutral account costs the share $\nu$ of a review, and a banned ID outside the universe costs $a$ plus $\tilde m$, the median margin of the innocents in the universe. A set of bans therefore costs
\begin{equation}
  \text{cost} = \sum_{TP}a + \sum_{FN}\ell_u + \sum_{FP}(a + m_u) + \sum_{\substack{\text{neutral}\\\text{acted}}} a\nu + \sum_{\text{unknown}}(a + \tilde m),
  \label{eq:cost}
\end{equation}
and the task scores
\[
  \text{Savings} = 100\operatorname{clip}_{[0,1]}\left(1 - \frac{\text{cost}}{\min(C_{\text{ban none}}, C_{\text{ban all}})}\right),
\]
where $C_{\text{ban none}} = \sum_{T}\ell_u$ and $C_{\text{ban all}} = \sum_{T}a + \sum_{I}(a + m_u)$. In the example of Appendix~\ref{app:examples}, $a$ is \$2 and $\nu$ is one half. This follows the example-dependent costs of cost-sensitive classification~\citep{elkan2001foundations}, and the same formula prices payout holds.

Three choices in Equation~\ref{eq:cost} shape how a ban is scored. First, a neutral account is left out of both baselines, so leaving it open is never a miss, while banning it still costs a review. If neutral bans were free, a run could ban the whole population and keep the targets it swept up; if they were charged as false positives, finding real fraud outside the task's scope would be punished. Second, an ID that is in no universe cannot be priced from the key. Charging it only $a$ would make an invented ID nearly free, so it is charged the median innocent margin as well. Third, in 18 of the 45 \texttt{cost\_set} expectations a wrong ban destroys only part of the innocent's margin, one half or one tenth, which models appeals and reinstatements and makes precision matter less relative to recall. Across the final tasks, 38 expectations use $a = \$2$ and $\nu = \tfrac12$, three use $a = \$25$ and $\nu = 1$, and four charge no review, so that only losses and margins count. The savings before clipping can be negative, which means the bans destroyed more value than either baseline, and the grader reports that value although the score stops at 0.

\paragraph{Data sources.} A \texttt{publish\_data\_sources} filing is the table behind a dashboard tile: its columns, and the rows the tile would show. Each task carries a contract that names the columns, their types, and which of them form the key; the contract is also the schema the agent is given. A publication missing a contract column scores 0. Otherwise the filed frame is right when it has exactly the key rows of the answer and every cell matches: numbers within the column's tolerance, integers exactly, dates to the day, and text after normalization. The score is 100 for a right frame and 0 for any other, since a tile with any wrong row cannot be used. The share of rows that were right is reported, but does not enter the score.

\paragraph{Tables.} A \texttt{file\_schedules} filing is a named table with key columns and rows, such as a reconciliation of receivables to the general ledger. The grader picks the schedule by name, matches rows on the key columns as normalized text, and counts a row correct when every numeric column of the answer is present and within tolerance. Other filed columns are ignored, so a schedule may carry its workings. The score is 100 times the correct rows over the answer's rows plus, in 13 of the 14 tasks, the filed rows the answer does not contain, so a reconciling item invented to force a tie counts against the schedule. A schedule filed twice under one name keeps the later copy.

\paragraph{Keyed values.} A \texttt{report\_metrics} filing reports a value per key, such as a figure for each month, and a \texttt{remediate\_payments} filing an amount per account or pay period. A key is correct when its value is within the larger of an absolute tolerance and a relative one. The score is 100 times the correct keys over all keys in the answer, so a wrong or missing key costs the same. In 5 of the 35 expectations, keys the answer does not contain are added to the denominator as well, for questions where reporting an extra key is itself an error.

\paragraph{ID sets.} For an \texttt{id\_set} expectation, the grader takes the filed IDs as a prediction of the answer's set and scores 100 times their F1. Of the 32 such expectations, 27 grade bans whose key is a list of accounts rather than a priced universe, and the rest grade holds, remediations, pay periods, and a discontinued promotion code. If the answer set is empty, the question has no right target: filing nothing scores 100 and filing anything scores 0. The one \texttt{binary} expectation uses the same comparison but scores 100 only for an F1 of exactly 1, because remediating the right accounts together with a wrong one is still a wrong decision. That expectation is also gated on another in its task: it scores only if every shortfall it remediates was first reported within tolerance. Filing no remediation is correct when nobody was shorted, but a run that reported a wrong shortfall and then filed nothing earns credit only if the shortfalls were reported correctly.

\paragraph{Allocations.} Nine tasks ask for a plan that moves a budget across cells such as a zone, a day type, and an hour band: five remove quest dollars, two remove courier hours, and two add courier hours where they prevent the most cancellations. The answer key holds each cell's capacity and its marginal value in the simulator, the deliveries lost per hour removed, the cancellations avoided per hour added, or the net value of a quest dollar. The loss of a plan is the sum over cells of the amount moved times that marginal. The plan scores
\[
  100\operatorname{clip}_{[0,1]}\left(\frac{L_{\text{uniform}} - L}{L_{\text{uniform}} - L_{\text{ref}}}\right) \times f,
\]
where $L_{\text{uniform}}$ is the loss of moving the same share in every cell, which needs no analysis, and $L_{\text{ref}}$ is the top of the scale. For the four tasks that remove or add courier hours, it is the loss of the best plan the authors reached from the warehouse alone, and the plan that is optimal on the true marginals is reported but not used, since reaching it would require information that the agent cannot obtain. For the five quest-cut tasks, $L_{\text{ref}}$ is the optimal plan itself, so a score of 100 means the agent recovered the best cut under the simulator's response model. The factor $f$ is 1 while the total moved is within 2\% of the budget and falls linearly to 0 at 15\% short or over. A cut or addition above a cell's cap is clipped to the cap, a row naming no cell is rejected, and a row at a coarser grain, such as a whole zone, is spread over its cells in proportion to their hours.

\paragraph{Written notes.} A \texttt{presence} expectation checks only that a filing of a kind was made. All 34 are \texttt{note} filings, in which the agent explains its work. In 32 tasks, the note is optional and does not enter the score. In the two reconciliation tasks, it is scored, with a weight of 0.5 against 3 for the table it explains.

\paragraph{Answer keys.} Of the \ntasks tasks, 128 take their key from a local builder, and 82 from truth queries.

%% file: sec/appendix/H_spider2.tex
\FloatBarrier
\section{The structure of Spider 2.0-Snow}
\label{app:spider2}

We measure Spider 2.0-Snow~\citep{lei2025spider}, the split whose leaderboard the introduction quotes, with 547 questions over 152 databases. All numbers come from the task manifests, DDL, gold SQL, and gold-table lists in the public repository at commit \texttt{cafb867},\footnote{\url{https://github.com/xlang-ai/Spider2}, August 12, 2026. The original 632-task release is commit \texttt{9615ebf}.} and \texttt{analysis/spider2/spider2\_shape.py} in the \ours release reproduces them.

\begin{figure}[!ht]
\centering
\includegraphics[width=\textwidth]{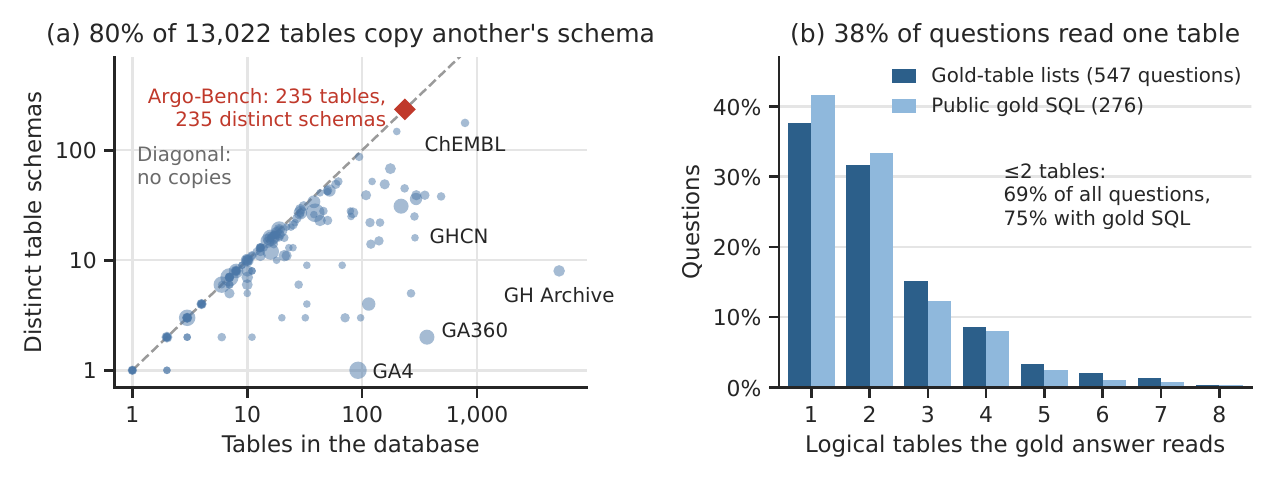}
\caption{Spider 2.0-Snow's databases are collections of shards, and its answers read few tables. (a)~Tables against distinct table schemas for each of the 152 databases, with point area proportional to questions. Distance below the diagonal is shards of one table. (b)~Logical tables read per question, with shards counted once, from the authors' gold-table lists (547 questions) and the public gold SQL (276).}
\label{fig:spider2}
\end{figure}

\begin{table}[!ht]
\caption{Source of each Spider 2.0-Snow question's database. Third-party BigQuery data includes ISB-CGC cancer genomics, Open Targets, and GH Archive.}
\label{tab:spider2-provenance}
\centering
\small
\begin{tabular}{@{}lrr@{}}
\toprule
Source of the database & Questions & Share \\
\midrule
BigQuery public data, Google & 341 & 62.3\% \\
BigQuery public data, third party & 39 & 7.1\% \\
Local SQLite files (Kaggle, textbook and vendor samples) & 135 & 24.7\% \\
Snowflake Marketplace & 18 & 3.3\% \\
Other public dumps (Meta Kaggle, WideWorldImporters, CDC) & 14 & 2.6\% \\
\bottomrule
\end{tabular}
\end{table}

\paragraph{Provenance.}
\label{app:spider2-provenance}
The Spider 2.0 paper splits its 632 tasks by host engine, with 214 on BigQuery and 198 on Snowflake. However, 180 of the Snowflake tasks use BigQuery public datasets copied into Snowflake, and only 18 use Snowflake Marketplace data. By source (Table~\ref{tab:spider2-provenance}), 69.5\% of Spider 2.0-Snow's questions run on BigQuery public data and 24.7\% on local SQLite files. A fifth (111 questions, 20.3\%) run on one of 22 synthetic, obfuscated, or textbook sample databases, such as Looker's synthetic theLook store and the obfuscated Google Analytics sample exports.

\paragraph{Shards.}
We count two tables in a database as one \emph{schema} when their column names and types are identical, as with the daily tables \texttt{ga\_sessions\_20160801} to \texttt{ga\_sessions\_20170801}. Of the 13,022 tables with a DDL, only 2,599 have a distinct schema, so 80.0\% copy another table (Figure~\ref{fig:spider2}a). Without \texttt{GITHUB\_REPOS\_DATE}, whose 4,989 identical daily tables hold a single event log, the share is 67.0\%. The median database has 11 distinct schemas, and the most varied has 177. The \ours warehouse has 235, with no copies. The median database has two questions, and 54 of the 152 have one.

\paragraph{Tables read.}
Counting shards once, 75.0\% of the 276 questions with public gold SQL read at most two logical tables, and 41.7\% read one. The authors' gold-table lists, which cover all 547 questions and agree with the gold SQL on 93.8\% of the questions both cover, give 69.3\% and 37.7\%, and no question reads more than eight (Figure~\ref{fig:spider2}b). The reference solution in Figure~\ref{fig:overview} reads 6 of the 235 \ours tables across three EBS modules.

\paragraph{Caveats.}
We skip 526 Cybersyn tables whose listings share no DDL. These counts describe the shape of the data, not question difficulty or the correctness of gold answers, which \citet{jin2026pervasive} audit.

%% file: sec/appendix/I_findings.tex
\FloatBarrier
\section{Findings in detail}
\label{app:findings}

Here, we provide the cases behind Section~\ref{sec:findings}, along with two fraud cases that compare prompt variants of one task. The numbers come from the 9,869 completed and graded runs over the \ntasks tasks at low, medium, high, and extra-high effort (47 settings of model and effort), and the traces of all of them. Scores are on the 0 to 100 scale of Table~\ref{tab:results}, except that a forecast which misses by more than a typical reference scores below 0 (Appendix~\ref{app:grading}). Paired comparisons hold the model and effort setting fixed, and each setting has one run per prompt, so differences between prompt variants are descriptive and not controlled experiments.

\FloatBarrier
\subsection{Confidence intervals}

\begin{table}[!ht]
\caption{Overall results with 95\% bootstrap confidence intervals over task families, each model at its setting in Table~\ref{tab:results}.}
\label{tab:ci}
\centering
\footnotesize\renewcommand{\arraystretch}{0.95}
\begin{tabular}{@{}lll@{}}
\toprule
Model & Score & Solved (\%) \\
\midrule
Claude Opus 5.5 & 59.5 [52.0, 67.1] & 34.8 [27.3, 43.3] \\
GPT-6 Astra & 51.8 [43.8, 59.8] & 27.6 [20.8, 35.3] \\
Claude Sonnet 5.5 & 51.8 [44.5, 59.0] & 28.6 [21.5, 36.5] \\
GPT-6.1 Sol & 49.5 [41.8, 57.0] & 24.8 [18.4, 32.0] \\
GPT-6 Sol & 36.8 [29.5, 44.0] & 17.6 [12.3, 23.8] \\
Kimi K3 & 28.4 [21.0, 36.7] & 14.3 [9.5, 19.7] \\
Gemini 3.8 Flash & 26.2 [20.4, 33.2] & 16.7 [11.5, 22.5] \\
DeepSeek V4.1 Flash & 25.4 [19.6, 34.7] & 17.6 [12.1, 23.7] \\
Qwen 3.8 Max & 24.4 [18.6, 31.4] & 17.1 [12.0, 23.1] \\
GLM 5.3 Flash & 21.2 [16.3, 28.4] & 12.4 [8.0, 17.4] \\
Muse Spark 1.3 & 20.7 [15.6, 26.9] & 13.8 [9.4, 18.8] \\
Claude Sonnet 5 & 17.3 [12.8, 22.7] & 9.5 [5.8, 13.9] \\
GPT-6 Luna & 15.0 [10.8, 19.6] & 7.6 [4.3, 11.6] \\
Claude Haiku 4.5 & 5.5 [3.4, 8.1] & 1.4 [0.0, 3.3] \\
\bottomrule
\end{tabular}
\end{table}

Table~\ref{tab:ci} gives 95\% bootstrap confidence intervals for the overall results of Table~\ref{tab:results}, from 10,000 resamples of the 146 task families rather than of the 210 tasks. A family is a base scenario together with its prompt variants, such as the hinted and unhinted versions of a task or the four search widths of the refund-collusion task, so variants of one scenario do not count as independent evidence. Families are named by their task stem, which also merges a few distinct scenarios and so errs toward wider intervals. Clustering widens each side of an interval by at most 1.6 points on the score and 1.9 on the solved rate over resampling tasks. Paired by task, with families resampled, Claude Opus 5.5 leads GPT-6 Astra by 7.7 points [1.4, 13.7], Claude Sonnet 5.5 by 7.7 [2.3, 13.4], and GPT-6.1 Sol by 10.0 [3.6, 16.3], while Astra, Sonnet 5.5, and GPT-6.1 Sol are not separated (Astra leads Sonnet 5.5 by 0.1 [$-$6.5, 6.6] and GPT-6.1 Sol by 2.3 [$-$1.2, 6.0], and Sonnet 5.5 leads GPT-6.1 Sol by 2.2 [$-$4.4, 8.7]). GPT-6.1 Sol leads its predecessor GPT-6 Sol by 12.8 [6.8, 19.1] and Kimi K3 by 21.2 [13.8, 28.5]. Astra leads GPT-6 Sol by 15.1 [9.2, 21.4], Sonnet 5.5 leads GPT-6 Sol by 15.0 [8.9, 21.1], and GPT-6 Sol leads Kimi K3 by 8.4 [0.5, 15.4]. Each setting has one run per task, so the intervals reflect which tasks are in the set and not the variance between runs of the same task.

\FloatBarrier
\subsection{Reasoning effort}

Table~\ref{tab:effort} gives every model's mean score at each reasoning effort it offers, on the tasks it completed at every such level. Effort helps the GPT-6 and Claude models at every step, with diminishing returns only for Opus and GPT-6.1 Sol, and the step from high to extra-high adds 1.9 to 11.5 points for them. Kimi K3 gains 10.1 points from low to high and 5.8 more at extra-high, and GLM 5.3 Flash 4.3 from high to extra-high. Past medium, Gemini 3.8 Flash (18.3, 26.9, 26.2) and Qwen 3.8 Max (21.1, 26.3, 24.4) score lower than at medium, Muse Spark 1.3 (16.7, 20.2, 19.7, 20.7) dips at high and recovers at extra-high, and DeepSeek V4.1 Flash scores 21.6 at low, 20.5 at high, and 25.4 at extra-high. Each of these drops is under 2 points and within the confidence intervals of Table~\ref{tab:ci}, so we do not read them as an effect of effort. Two mechanisms are consistent with them. All 65 runs that exhausted the context window are from Qwen 3.8 Max (Appendix~\ref{app:tools}), whose window is shorter than the other models', and longer reasoning at higher effort fills it sooner. For the other models, we suspect that runs with many tool calls, each returning data, bury the task's objective in a long context, and 43 of the 56 runs that reached the turn limit are from Muse Spark 1.3.

\begin{table}[!ht]
\caption{Mean score by reasoning effort, each model on the tasks it completed at every level it offers. A dash marks a level the model does not offer: Kimi K3 and DeepSeek V4.1 Flash map medium onto high, GLM 5.3 Flash maps low and medium onto high, Gemini 3.8 Flash and Qwen 3.8 Max map extra-high onto high, and Claude Haiku 4.5 has no effort setting (its single run is shown under high). For Kimi K3, DeepSeek V4.1 Flash, and GLM 5.3 Flash, extra-high is the vendor's maximum setting.}
\label{tab:effort}
\centering
\small
\begin{tabular}{@{}lrcccc@{}}
\toprule
Model & Tasks & Low & Medium & High & Extra-high \\
\midrule
Claude Opus 5.5 & 210 & 32.6 & 49.6 & 55.5 & 59.5 \\
GPT-6 Astra & 210 & 33.3 & 45.1 & 47.6 & 51.8 \\
Claude Sonnet 5.5 & 210 & 19.5 & 25.4 & 40.3 & 51.8 \\
GPT-6.1 Sol & 210 & 24.6 & 35.9 & 43.6 & 49.5 \\
GPT-6 Sol & 210 & 19.5 & 28.2 & 30.7 & 36.8 \\
Kimi K3 & 210 & 12.5 & -- & 22.6 & 28.4 \\
Gemini 3.8 Flash & 210 & 18.3 & 26.9 & 26.2 & -- \\
DeepSeek V4.1 Flash & 210 & 21.6 & -- & 20.5 & 25.4 \\
Qwen 3.8 Max & 210 & 21.1 & 26.3 & 24.4 & -- \\
GLM 5.3 Flash & 210 & -- & -- & 16.9 & 21.2 \\
Muse Spark 1.3 & 210 & 16.7 & 20.2 & 19.7 & 20.7 \\
Claude Sonnet 5 & 209 & 11.3 & 11.7 & 14.2 & 17.4 \\
GPT-6 Luna & 210 & 5.7 & 12.3 & 13.1 & 15.0 \\
Claude Haiku 4.5 & 210 & -- & -- & 5.5 & -- \\
\bottomrule
\end{tabular}
\end{table}

\FloatBarrier
\subsection{Wrong record}

The task behind the marketing dashboard of Section~\ref{sec:findings} covers 11 campaigns run in April and May 2024. Each campaign is a push notification sent in a few versions, one carrying each of the campaign's discount offers and one with no coupon attached, the reminder. Every notification sent is one customer-window, a customer and the response period that follows it. For each offer, and separately for members and non-members, the dashboard reports how many windows the offer was sent to, how many the reminder was sent to, the booked order contribution per 1,000 windows of each, and the difference between the two. Whether a customer counts as a member is decided at the moment the message was sent, trials and comped memberships included, and the prompt says that effective starts are inclusive and ends exclusive. That gives 62 rows. Table~\ref{tab:record} sorts the 47 settings by what they published.

\begin{table}[!ht]
\caption{The 47 settings on the marketing-dashboard task, by what they published. The last column is how many more offer windows the run counted as members' than the key does (861,125).}
\label{tab:record}
\centering
\small
\begin{tabular}{@{}lrr@{}}
\toprule
Published & Settings & Member windows over the key \\
\midrule
The right table & 3 & 0 \\
Right rows, membership from contract dates cut at termination & 12 & $+$7,664 \\
Right rows, membership from contract dates uncut & 12 & $+$36,824 \\
Right rows, other errors & 6 & \\
Wrong key column & 4 & \\
Nothing & 10 & \\
\bottomrule
\end{tabular}
\end{table}

The three right tables come from Claude Opus 5.5, GPT-6 Astra, and GPT-6.1 Sol at extra-high effort, and the same three models fail at low, medium, and high effort. Most of the work is right in many more runs. Of the 30 that publish the right rows, 28 count how many windows each offer was sent to correctly in total, and ten get every offer's booked contribution within \$2.30 of the key, on totals of up to \$327,727: the three passing runs, Astra at medium and high, Opus at low and medium, GPT-6.1 Sol at medium and high, and Gemini 3.8 Flash at high effort. The seven of those that fail get one thing wrong, which windows belong to members. The four runs with the wrong key column label each offer by its number within the campaign instead of its identifier.

What goes wrong is where the runs look for membership. A membership lives in the warehouse twice. It is a service contract with a start date, an end date, and, once cancelled, a termination date, and it is a billing history of timestamped events: trial started, converted, renewed, charge declined, past due, retry, reactivated, cancelled. The contract line also states a grace period of seven days, the same on all 2,507,271 lines. When a member's charge is declined, the benefits continue through those seven days and then stop until a retry succeeds. If none does, the contract is cancelled about three weeks after the decline, and only then is its termination date written, so for the two weeks in between the contract looks live and the customer is no longer a member. Twelve runs call a customer a member whenever a contract's dates contain the time of the message, cut off at the termination date, and twelve more use the dates without the cut. Against the key, which follows the billing history, the first rule counts 8,379 offer windows as members' that were not: 5,832 sent while the customer was past due beyond the grace, 1,757 sent on the day the contract was created but before the hour it was, since the dates carry no time of day, and 790 sent on the day of a cancellation but after it. It also misses 715 real members, so it ends up 7,664 windows over. The second rule further counts 28,745 windows on contracts that had been cancelled but had not reached their scheduled end date, and a few hundred more at the margins, and ends up 36,824 over. These are small shares of the 861,125 member windows, but they fall on the member side, where some cells hold only a few hundred windows. Under the first rule the member cells move by a median of 1.1\% of their windows and by up to 27\%, their difference per 1,000 windows moves by a median of \$8.6 and by up to \$3,610, and one of the 31 changes sign.

The three passing runs build membership from the billing history and then check it against a table the warehouse also has, the orders on which a member benefit was actually applied. Opus lines up orders by how many days had passed since the customer's last declined charge: every order in the first seven days carries a benefit, and none from the eighth day on, which is the grace period read straight off the data. Its final reconstruction misses no member and adds 102 among 4.9 million April and May orders. Astra assumes the seven days, finds that without them 1,823 orders on two sample days carry a benefit its reconstruction denies and with them none, and then confirms the seven days on the contract line. GPT-6.1 Sol takes the longest road. It first defines a member as anyone whose contract runs from creation to cancellation, the rule of twelve failing runs, and files a note saying so. A check on the 198,230 orders of April 1 finds 1,057 that this rule calls members' and that received no benefit. It rebuilds from the billing events but without the grace, and the error now runs the other way, with 903 orders that carry a benefit its rule denies. It reads the seven-day term, rebuilds once more, is left with seven mismatches, and files a note that supersedes the first.

The failing runs had the same tables, and some of them ran the same check. Claude Sonnet 5.5 at high effort compares its rule with the benefits on the orders of May 10, finds 1,363 inside the contract dates that got no benefit, lists them, and sees that they belong to contracts in payment failure, terminated days later. It then tries two rules, one that ends benefits at the declined charge, which 976 orders contradict, and one that keeps them until cancellation, which 1,247 contradict, chooses the second, and notes that ``PAST\_DUE dunning suspension is not treated as loss of entitlement''. Opus at high effort checks in one direction only, that 99.9\% of the orders with a benefit fall inside its intervals, which is true and cannot catch a member counted after the grace, and files. Gemini 3.8 Flash at high effort finds 33,032 benefit rows outside their contract's dates and moves on.

\FloatBarrier
\subsection{Wrong objective}

The quest-cut tasks ask where to remove \$9.6 million of the \$24.0 million paid in quest bonuses in 2024. The grader scores a plan by the surge pay the simulator adds back when quests are removed. Table~\ref{tab:questcut} contrasts two plans on the version without hints. Both meet the budget, and a grader that checked only feasibility would pass both.

\begin{table}[!ht]
\caption{Two plans for the quest-cut task without hints, both at extra-high effort, evaluated under the simulator's frozen response model.}
\label{tab:questcut}
\centering
\small
\begin{tabular}{@{}lrr@{}}
\toprule
 & GPT-6 Astra & Claude Opus 5.5 \\
\midrule
Score & 0 & 99 \\
Budget met & yes & yes \\
Net saving & $-$\$86,281 & \$3,089,763 \\
Saving of a uniform cut & \$402,037 & \$402,037 \\
\bottomrule
\end{tabular}
\end{table}

Astra finds the random holdout, estimates local and citywide supply responses with fixed effects and partial pooling, and allocates cuts by the courier-hours and queue pressure it expects to lose. Opus finds that launched quests have little measured effect on service but substitute for surge pay, and cuts the cells with the least surge reduction per bonus dollar. The optimum under the response model saves \$3.116 million. When the prompt states the purpose of quests and the existence of the holdout, Astra changes its decision rule and scores 74.9. Over 47 matched settings, this prompt raises the mean score from 18.5 to 61.7, with 41 settings improving, three worsening, and three tied. At weekly grain, adding the purpose, the holdout, and the economics of the pay floor raises the mean from 8.2 to 39.4 over 47 settings. These are bundled changes to the prompt and not isolated tests of any one hint. The savings are evaluations under the frozen response model, not outcomes of a live deployment.

\FloatBarrier
\subsection{Wrong quantity}

\paragraph{Location reports.} The courier app sends a location report on every job event and, while a courier is online without a job, every half hour. The warehouse keeps the event-triggered reports but only hourly idle check-ins. Of the 47 completed runs, 29 return exactly 13,575,751 reports for January against 14,541,453 in the key, which is the retained count. GPT-6 Astra checks the minute and second distribution, recognizes the mismatch, and writes that its count does not establish the volume the app sent, and then files it. Claude Opus 5.5 reconstructs the reports from sessions and assignments over 86 SQL and 32 Python calls, but counts order assignments rather than app events and reports 21,617,842, 48.7\% above the key. Counting the retained events and adding the retained idle check-ins once more lands within the 1\% tolerance. Claude Sonnet 5.5 at extra-high effort is the one run that passes: it finds that idle check-ins are stored only on the hour and adds back the half-hour ones, sizing them from gaps in the report IDs and checking the result against a second count of the hourly check-ins.

\paragraph{Kitchen capacity.} Asked for orders turned away because kitchens were full, 27 of 47 runs return 12,580 orders at 3,550 storefronts for January, and the other 20 return 1,531 orders at 986 storefronts. The key is 5,811 orders at 516 storefronts. At high effort, Claude Opus 5.5 finds the platform's capacity refusal code and explicitly rejects it, writing that it is ``deliberately NOT using'' the field, in favor of the reason merchants enter when they reject an order. Stating that the event is the platform refusing the order at admission, which is distinct from a merchant rejecting it on the tablet, lifts passes from 0 of 47 to 18 of 47. The underspecified prompt contributes to this failure, and we report the pair for that reason. A related task on the readiness of point-of-sale systems is failed by all 47 completed runs.

\paragraph{Payout timing.} Only 9 of 47 completed runs pass a dashboard task on the dollar-days that merchant payouts spend withheld. Astra and Opus agree on the year-end withheld balance (\$800,670.93), the cumulative amount received (\$6,397,902.81), and zero in transit. Astra dates payments by the payment date and scores 100. Opus dates them by the date they cleared the bank, stating that it ignored the payment dates, and scores 0. Its lower bound on withheld dollar-days is 206,139,983.33 against Astra's 192,119,146.62, a difference of 7.3\%. The basis changes the daily balances from the first week of July and reorders the top 20 invoices below rank 9, replacing one of them, while leaving every year-end total unchanged. Across all dashboard tasks, 1,171 of 1,750 data sources that pass their structural contract (66.9\%) score zero on their values. These are 1,879 graded sources from 1,691 runs on 36 tasks, since some tasks publish several sources.

\FloatBarrier
\subsection{Past labels and benign lookalikes in fraud}

\paragraph{Historical labels.} One task asks the agent to review the 308 couriers the risk engine first warned in the second half of the year and never deactivated, and to deactivate those who steal orders. The longer prompt describes the scheme and its economics and says that what the engine did with the couriers it warned in the first half shows how a thief and an unlucky courier each look. GPT-6 Astra at high effort takes that history as its labels, counting a first-half courier as a thief if the engine deactivated it or the desk confirmed a theft case. It builds logistic regression, random forest, gradient boosting, and beta-binomial models, cross-validates them, adds route features, and reaches a temporal-holdout AUC of .932 against those labels. The history is reliable in one direction only. Of the 659 couriers warned in the first half, the engine deactivated 203, of whom 199 are thieves, but 254 of the 456 it never deactivated are thieves too. The couriers Astra treats as honest therefore include many thieves the engine missed, and the second-half queue consists, by construction, of couriers the engine has not caught. After applying estimated economic thresholds and further eligibility rules, Astra acts on one courier, who is outside the target set, and catches none of the 78 targets. Its filed note calls the labels proxies, not independent proof of intent. The shorter version of the prompt drops the scheme, its economics, and the pointer to the engine's history. On it, the same setting checks physical delivery evidence against destination buildings, files 65 deactivations, and catches 40 of the 78 targets with no innocent courier banned and 25 neutral actions (score 59.2). Over 47 matched settings, the shorter prompt raises the mean score from 8.8 to 32.0, with 37 settings improving, six worsening, and four tied. The evidence that separates thieves from unlucky couriers is in the warehouse, but the longer prompt recommends the labels Astra used and differs from the shorter one in three places, so the pair shows how prompt wording steers models rather than a model choosing poor labels unprompted. It does not show that classifiers are worse than rules.

\paragraph{Benign lookalikes.} The first three variants of the refund-collusion family keep the story and widen the population of storefronts to search. Table~\ref{tab:col22} gives the mean over the 47 settings that completed all four variants. The target set also grows from eight to ten identifiable rings, so this is a comparison of scope and not a pure test of data volume.

\begin{table}[!ht]
\caption{Mean score on the refund-collusion family by search population, over 47 matched settings.}
\label{tab:col22}
\centering
\small
\begin{tabular}{@{}lr@{}}
\toprule
Storefronts searched & Mean score \\
\midrule
1,500 to 15,000 orders & 28.6 \\
1,500 to 45,000 orders & 16.8 \\
Every storefront & 21.6 \\
Every storefront, with fewer hints about the mechanism & 15.5 \\
\bottomrule
\end{tabular}
\end{table}

On the full search with hints, Claude Opus 5.5 at extra-high effort scores 88.9 after 136 Python calls and 60 SQL queries. It infers each storefront's liability tier from the share of its refunds the storefront pays, finds missing-item refunds that charge a different share at 98 storefronts, and then splits them. At 72, three to five regulars each claim on their second and third orders and keep ordering without another claim. These are the task's benign control, a sloppy kitchen with loyal regulars, and Opus leaves them alone. At the other 26, a group of regulars claims on roughly one order in three all year. Opus freezes the payouts of all 26, the ten target rings and 16 other fraudulent storefronts that the key leaves ungraded, and so of no innocent storefront, and its 130 customer closures include 70 of the 80 targets. At high effort, Opus finds the same 98 storefronts, stops there, and freezes the payouts of all of them (score 9.4). Most runs fail in a similar fashion: of the 16 Opus and Astra runs on the two full searches, 11 freeze all ten rings together with 50 to 71 innocent storefronts, nearly all of them benign controls. GPT-6 Astra holds the ten rings and no innocent storefront on the full search with hints at low and high effort (88.6 and 90.6), but not at medium and extra-high effort, so one run per setting says little about a model's reliability on this task. On the version with fewer hints, none of the eight Opus and Astra runs does; the best, Opus at low effort (69.9), holds eight of the ten rings and no innocent storefront. The 98 storefronts are exactly those whose refunds the simulator injected, which shortcuts the search (Appendix~\ref{app:audit}), so we do not present the successful runs as evidence that the approach would transfer to real fraud.

\FloatBarrier
\subsection{Forecasts}

Across 4,553 forecast series filed in 3,346 runs on 72 tasks, nominal 80\% intervals contain the realized value 2,040 times (44.8\%). Weighting tasks equally gives 46.0\%. Coverage is 43.8\% on series scored against a no-change reference and 48.2\% on series scored against a fixed reference. The references set each series' scale rather than compete as forecasts, and apart from its floor the grade is proper however well they are calibrated. Even so, the no-change reference's own 80\% interval contains the outcome in 21 of the 32 series outside December (66\%), where the models' intervals contain it 44.7\% of the time. A no-change forecast cannot anticipate the holiday rise by construction, and with only 2024 in the warehouse no method can learn it from history: on the 44 December series it covers 15, and models cover 43.2\%. The fixed references are mostly figures the prompt asks the agent to check, so their coverage (10 of 22) reflects the task design rather than how forecastable the series are. Of the 4,553 series, 1,691 grade below zero, further from the outcome than the median reference; many of them are complete forecasts of the right order of magnitude. These series come from one simulated world and a selected set of tasks, so they are correlated and not independent calibration trials.

\paragraph{Incentives of the grade.} We checked on these filings whether the grade of Appendix~\ref{app:grading} rewards honest forecasts. Treating each filing as its forecaster's belief, a split normal whose 10, 50, and 90\% quantiles are the filed interval and point, we searched reports that move the point part or all of the way to the reference and scale the interval by 0.25 to 3, and compared their expected grades under that belief. For 96.7\% of the 4,527 series with a well-formed 80\% interval, filing the belief itself maximizes the expected grade; in the other 3.3\%, the $-200$ floor makes a narrower interval (in a few series, a point moved toward the reference) pay, by 3.6 points on average there and 0.12 across all series. The floor binds on 464 of the 4,606 forecast expectations (10.1\%). The per-series skill score $100\max(0, 1 - \text{WIS}/\text{WIS}_{\text{ref}})$, with the reference's WIS taken at the outcome, is a common alternative; under it the honest filing would be the best one for only 18.6\% of series, a narrower interval would pay in 71.9\% and moving toward the reference in 28.8\%. Dividing by the reference's WIS at the outcome weights each series by how well its reference happened to do, and the floor at zero makes extra risk free whenever skill is likely to be negative. A fixed denominator on the original scale, such as the reference's expected WIS, would be proper, but because the references are themselves overconfident it lets a handful of large misses dominate the mean, which motivates the logarithmic tail of the asinh scale.

\paragraph{Dependence on the reference.} The reference still sets the scale of each series, so a tighter reference lowers the grade of the same forecast. Table~\ref{tab:rescore} grades the June base-pay forecasts against the reference of their harder variant, FP\&A's plan, holding the points, intervals, and outcomes unchanged. The forecasts have a median absolute percentage error of 1.57\%, and their mean grade falls from 84.7 to 6.0, since the plan's standard deviation is a thirteenth of the no-change forecast's (\$1.0 million against \$13.2 million). On the harder variant itself, 32 of 47 point forecasts are within 3\% of the truth and 46 of 47 intervals cover it, yet the mean grade is 4.8, because the median interval is nearly three times as wide as the plan's. We therefore report interval coverage and absolute error alongside the grade.

\begin{table}[!ht]
\caption{The June base-pay forecasts graded against the fixed reference of their harder variant, with points, intervals, and outcomes held fixed.}
\label{tab:rescore}
\centering
\small
\begin{tabular}{@{}lrrr@{}}
\toprule
Task & Runs & No-change reference & FP\&A's plan \\
\midrule
June courier base pay & 47 & 84.7 & 6.0 \\
\bottomrule
\end{tabular}
\end{table}

\paragraph{References set with the outcome in view.} Seven of the 22 fixed references (Table~\ref{tab:fixed-refs}) were set with the realized value in view: five at 2.4 to 3\% above it, one from the simulator's label, and one from losses reported after the cutoff. Each is a figure the prompt attributes to the business and asks the agent to check, so the prompt carries it too. Under this grade the reference only centers and scales its series and does not set the zero, so filing the figure unchanged grades about 30 on the four set at 1.024 to 1.026 times the realized value, and 60 on Finance's surge read, rather than 0. Without the six tasks that carry them, no setting's score moves by more than 3.8 points. At the settings of Table~\ref{tab:results}, Claude Opus 5.5 stays first (60.7), and the order changes only between models less than a point apart: Claude Sonnet 5.5 passes GPT-6 Astra (53.2 and 52.3), and DeepSeek V4.1 Flash passes Gemini 3.8 Flash (27.1 and 27.0).

%% file: sec/appendix/J_task_audit.tex
\FloatBarrier
\section{Task audit}
\label{app:audit}

Because answer keys are computed from the latent state, a questionable key or prompt is a bug that can be traced and fixed. We list the open issues found while reading the traces and how we treat them.

\paragraph{Scope of identity substitution.} The prompt asks for accounts to take off the platform, but the original key also named couriers the platform had already deactivated, 138 of the 249 witnessed targets and 130 of those for an identity mismatch, so agents that left them out were penalized. We re-keyed the task to the 111 couriers still active at the end of 2024 and regraded every run, and it counts toward the headline numbers. On the new key the task is hard at every setting: over 47 settings the mean score is 10.8 without the description of the mechanism and 8.4 with it, and the best run scores 69.4 (Claude Opus 5.5 at extra-high effort).

\paragraph{Composite score in the refund-collusion family.} The prompt described equal costs for a missed ring, a wrongly held storefront, and a wrongly closed customer account, while the grader combines the storefront savings (weight 2) with the F1 score of customer closures (weight 1). We have aligned the prompt with the grader: it now states that the storefronts are two thirds of the review and how customer closures are judged. We then ran all four variants again on every setting with the new wording, and every score in the paper, including the overall results of Table~\ref{tab:results}, uses these runs. In Appendix~\ref{app:findings}, we discuss component counts rather than the composite.

\paragraph{Injected refunds in the refund-collusion family.} Every organic missing-item refund charges the storefront 0, 25, 50, 75, or 100\% of the item, according to its liability tier. The simulator instead splits the refunds it injects for the rings and for their benign control around a target share with random noise, so 664 of the 424,807 missing-item refunds of 2024 charge the storefront a share that matches no tier. All 664 fall at the 98 storefronts the simulator injected: the 20 phantom-item ring storefronts, the 72 benign controls, and six storefronts in three closed-and-reopened pairs. One pass over the refunds therefore finds every candidate, and 13 of the 16 Claude Opus 5.5 and GPT-6 Astra runs on the two full searches freeze payouts only within this set, eight of them at all 98. The family therefore tests whether an agent can tell a ring from its benign control, not whether it can find either among every storefront, and the widening in Table~\ref{tab:col22} tests less than it was designed to. Correcting the split changes the world, so we keep the family as run.

%% file: sec/appendix/K_warehouse_cost.tex
\FloatBarrier
\section{Warehouse cost}
\label{app:bigquery}

The cost column of Table~\ref{tab:results} counts model API spend only. The warehouse was a second cost of the same order. We ran the experiments on BigQuery, which bills a query by the bytes it scans, and Table~\ref{tab:bigquery} charges each run for the bytes its \texttt{run\_sql} queries scanned at the on-demand list price of \$6.25 per TiB. Across the 9,869 graded runs of the final experiment at every effort level, the agents scanned 986~TiB, or \$6,160 of warehouse compared to \$14,502 of model API spend. A query could scan at most 20~GiB (Appendix~\ref{app:tools}), and the number of queries was not limited.

\begin{table}[!ht]
\caption{BigQuery spend per task, each model at its setting in Table~\ref{tab:results}, over the same 210 runs per model. Mean and median are dollars per task at \$6.25 per TiB scanned. Share of cost is the warehouse's share of the run's total cost (model API plus BigQuery). Unsolved is the share of the model's BigQuery dollars spent on tasks whose score was below 5 out of 100, and per solve is its BigQuery dollars divided by the number of tasks it solved completely (a task score of at least 95). Rows are ordered by score.}
\label{tab:bigquery}
\centering
\small
\setlength{\tabcolsep}{4pt}
\begin{tabular}{@{}lrrrrrr@{}}
\toprule
 & \multicolumn{2}{c}{BigQuery \$ per task} & & & & \\
\cmidrule(lr){2-3}
Model & Mean & Median & TiB per task & Share of cost (\%) & Unsolved (\%) & \$ per solve \\
\midrule
Claude Opus 5.5 & 0.82 & 0.27 & 0.132 & 15 & 9 & 2.37 \\
GPT-6 Astra & 0.61 & 0.13 & 0.097 & 18 & 15 & 2.20 \\
Claude Sonnet 5.5 & 0.68 & 0.28 & 0.109 & 15 & 30 & 2.39 \\
GPT-6.1 Sol & 0.60 & 0.16 & 0.096 & 55 & 22 & 2.43 \\
GPT-6 Sol & 0.80 & 0.18 & 0.128 & 49 & 42 & 4.53 \\
Kimi K3 & 0.83 & 0.20 & 0.132 & 15 & 56 & 5.79 \\
Gemini 3.8 Flash & 0.86 & 0.33 & 0.137 & 18 & 41 & 5.16 \\
DeepSeek V4.1 Flash & 1.31 & 0.42 & 0.209 & 76 & 47 & 7.42 \\
Qwen 3.8 Max & 0.54 & 0.23 & 0.086 & 16 & 54 & 3.14 \\
GLM 5.3 Flash & 0.94 & 0.21 & 0.151 & 73 & 51 & 7.63 \\
Muse Spark 1.3 & 1.63 & 0.68 & 0.260 & 25 & 57 & 11.77 \\
Claude Sonnet 5 & 0.46 & 0.09 & 0.073 & 18 & 59 & 4.81 \\
GPT-6 Luna & 0.65 & 0.13 & 0.103 & 92 & 69 & 8.47 \\
Claude Haiku 4.5 & 0.25 & 0.03 & 0.040 & 56 & 84 & 17.55 \\
\bottomrule
\end{tabular}
\end{table}

The warehouse bill does not follow the model bill. The cheaper models spend proportionally far more on BigQuery. It is 92\% of the cost of a GPT-6 Luna run and over 70\% for GLM 5.3 Flash and DeepSeek V4.1 Flash, against 15\% for Claude Opus 5.5, so a Luna task costs \$0.70 all in rather than the \$0.06 of Table~\ref{tab:results}, and the gap between the cheapest and the most expensive models is far narrower than the API column suggests. Some also spend more in absolute terms. DeepSeek V4.1 Flash, GLM 5.3 Flash, and Muse Spark 1.3 spend \$0.94 to \$1.63 per task on the warehouse, against \$0.82 for Opus, because they query more and solve less. Muse Spark 1.3 scans 1.2 times as much as the next model, 55~TiB over its 210 tasks, and spends \$11.77 of warehouse for each task it solves completely, against \$2.37 for Opus and \$2.20 for GPT-6 Astra. Gemini 3.8 Flash spreads its scans over 197 model calls of mostly small queries of 0.7~GB each and spends 41\% of them on tasks that score below 5 out of 100, while Opus and Astra spend 56\% and 54\% of their warehouse dollars on tasks they solve completely. Spend is also skewed. Every median is below its mean, from \$0.03 to \$0.68, because a minority of runs scan close to the 20~GiB cap query after query, up to \$11.36 for a single Muse Spark 1.3 run.

%% file: sec/appendix/L_example_tasks.tex
\section{Example tasks}
\label{app:examples}

We walk through three tasks end to end: the forecasting task of Figure~\ref{fig:overview}
(Table~\ref{tab:example-forecast}), a dashboard data source (Table~\ref{tab:example-dashboard}),
and a two-sided fraud task (Table~\ref{tab:example-collusion}). Each card shows the prompt as the
agent saw it, the grader's scoring rule as a short Python sketch, and the reference solution, one
step per tool call (Appendix~\ref{app:tools}). The listings are the released reference files with
print statements and filed notes removed and long lines wrapped. Warehouse tables and Mission
Control calls are shown in green. Each card ends with the reference solution's score and each model's best score.

\clearpage
\cardcaption{tab:example-forecast}{A forecasting task after the April minimum-pay shock.
  The agent reads the warehouse as it stood on April 30 and is told the dispatcher's plan
  for May; the key is May's actual value, which the agent never sees.}

\cardbegin{fc-12-true-ups-may}{Forecasting}{warehouse as of April 30, 2024 $\cdot$ horizon May 2024}

\cardsection{Question}
\begin{cardprompt}
In April, the platform started to comply with the DCWP minimum-pay rule for food delivery
workers (Q1 was a tolerated phase-in period), and the new minimum of \$19.56 per hour came
into force. The minimum is owed on the time a courier is online, not just the time spent on
deliveries, and whenever a courier's pay for a week falls short of it, the platform tops it
up with a true-up. We bled hard on true-ups in the second half of April. On the same day our
platform updated the dispatcher to restrict when couriers can go online to the hours we
expect to need them, so the online time we pay for is busier, and it has kept tightening
since. For the payout periods ending in May we now expect about 2.31 million connected hours
in total.

Based on what you know, the numbers from January to April and that plan, file a
\code{minimum\_pay\_true\_ups\_usd} forecast for the sum of the ADJUSTMENTS lines alone on
the COURIER\_WEEKLY pay invoices for the payout periods ending in May, with an 80\% central
prediction interval.

Pay is the sum of base, incentives and adjustments on the weekly courier AP invoices for
the periods. Tips are not pay! End date from XX\_PAYOUT\_PERIODS falling in the month is
how we track whether a payout period falls in the month. A shift's online (connected) time
is the window between its start date and end date, and a shift belongs to the period its
start date falls in.
\end{cardprompt}

\cardsection{Grader}
{\footnotesize
The key is May's actual value, read from the full-year warehouse the agent never sees: the ADJUSTMENTS lines on the COURIER\_WEEKLY invoices for the four payout periods ending in May sum to \$33,266. The filing is scored by its weighted interval score
(WIS; \citealp{bracher2021evaluating}): the absolute error of the point averaged with the interval
score of \citet{gneiting2007strictly}, which charges width at par and a miss at ten dollars
per dollar. The scale is set by the no-change forecast built from the four months the agent
could see, April repeated with the interval of \citet[\S5.5]{hyndman2014forecasting}: the filed
quantiles and the outcome are mapped to $\operatorname{asinh}((v - r)/s)$, with $r$ and $s$ that forecast's point and standard deviation, and the grade is $100(1 - \text{WIS}/0.623)$ on that scale, where 0.623 is the median of the references' WIS across the benchmark (Appendix~\ref{app:grading}). The no-change forecast itself grades $-1$ here, since it misses May by about as much as the median reference misses. Even a forecaster who knew May's pay level exactly would miss a little, because the week of April 29 ran \$124k short and its two visible days misjudge it by about 8\%.\par
\cardlabel{SCORING}{}
\begin{lstlisting}[style=cardpy,numbers=none,xleftmargin=6pt,framexleftmargin=6pt]
def wis(point, lower, upper, y, level=0.8):       # Bracher et al. (2021), one interval
    alpha = 1 - level
    miss = 2 / alpha * (max(lower - y, 0) + max(y - upper, 0))
    return (abs(y - point) / 2 + alpha / 2 * (upper - lower + miss)) / 1.5

r, s = no_change.point, no_change.sd              # April, and sigma * sqrt(h)
g = lambda v: asinh((v - r) / s)                  # linear near r, logarithmic beyond
score = wis(*map(g, filed), y=g(truth))           # 0.036 for the reference solution
grade = 100 * max(-2, 1 - score / 0.623)          # 94; the no-change forecast grades -1
\end{lstlisting}
\vspace{2pt}
\vspace{4pt}
\centering
\setlength{\tabcolsep}{3.5pt}
\begin{tabular}{@{}l r r r r >{\columncolor{keyfill}}r@{}}
\toprule
\code{minimum\_pay\_true\_ups\_usd} & Jan & Feb & Mar & Apr & May (key) \\
\midrule
ADJUSTMENTS, periods ending in the month & $-145{,}658$ & $-145{,}945$ & $-174{,}435$ & $1{,}443{,}893$ & $33{,}266$ \\
\bottomrule
\end{tabular}\par
\vspace{2pt}
\begin{tabular}{@{}l r r r r r@{}}
\toprule
 & point & 80\% interval & WIS & WIS, asinh scale & grade \\
\midrule
No-change forecast & $1{,}443{,}893$ & $[246{,}300,\ 2{,}641{,}486]$ & $771{,}910$ & 0.631 & $-1$ \\
Reference solution, overleaf & $-92{,}494$ & $[-92{,}494,\ 211{,}349]$ & $62{,}176$ & 0.036 & \textbf{94} \\
\bottomrule
\end{tabular}\par}
\chiprow{%
  \chipref{Reference solution}{94}%
  \chip{chipgood}{Claude Opus 5.5}{72}%
  \chip{chippart}{GPT-6 Sol}{5}%
  \chip{chipzero}{GPT-6.1 Sol}{$-$14}%
  \chip{chipzero}{GPT-6 Astra}{$-$17}%
  \chip{chipzero}{Claude Sonnet 5.5}{$-$25}%
  \chip{chipzero}{Qwen 3.8 Max}{$-$43}%
  \chip{chipzero}{Kimi K3}{$-$51}%
  \chip{chipzero}{Muse Spark 1.3}{$-$110}%
  \chip{chipzero}{Gemini 3.8 Flash}{$-$143}%
  \chip{chipzero}{DeepSeek V4.1 Flash}{$-$143}%
  \chip{chipzero}{GPT-6 Luna}{$-$189}%
  \chip{chipzero}{Claude Sonnet 5}{$-$200}%
  \chip{chipzero}{Claude Haiku 4.5}{$-$200}%
  \chip{chipzero}{GLM 5.3 Flash}{$-$200}}

\clearpage
\cardsection{Reference solution}
\cardstep{1}{run\_sql}{One row per payout week so far: what the minimum-pay engine tested and
  topped up, orders, shift hours and the invoiced ADJUSTMENTS. The last row is the open week.}
\noindent\begin{minipage}[t]{0.49\linewidth}
\begin{lstlisting}[style=cardsql]
WITH weeks AS (
  SELECT PAYOUT_PERIOD_ID,
         DATE(PERIOD_START_DATE) AS week_start,
         DATE(PERIOD_END_DATE) AS week_end
  FROM XX_PAYOUT_PERIODS
  WHERE PARTY_TYPE_CODE = 'DRIVER'
    AND PERIOD_START_DATE <= '2024-04-30'
),
orders AS (
  SELECT w.PAYOUT_PERIOD_ID, COUNT(*) AS orders,
    COUNTIF(EXTRACT(DAYOFWEEK FROM o.ORDERED_DATE)
            IN (2, 3)) AS orders_mon_tue
  FROM weeks w
  JOIN OE_ORDER_HEADERS_ALL o
    ON DATE(o.ORDERED_DATE)
       BETWEEN w.week_start AND w.week_end
  GROUP BY 1
),
shifts AS (
  SELECT w.PAYOUT_PERIOD_ID,
         SUM(DATETIME_DIFF(s.END_DATE, s.START_DATE,
             SECOND)) / 3600 AS shift_hours
  FROM weeks w
  JOIN XX_DRIVER_SHIFTS s
    ON DATE(s.START_DATE)
       BETWEEN w.week_start AND w.week_end
  GROUP BY 1
),
\end{lstlisting}
\end{minipage}\hfill
\begin{minipage}[t]{0.49\linewidth}
\begin{lstlisting}[style=cardsql,firstnumber=30]
invoices AS (
  SELECT CAST(i.ATTRIBUTE1 AS INT64)
           AS PAYOUT_PERIOD_ID,
         SUM(l.AMOUNT) AS adjustments
  FROM AP_INVOICES_ALL i
  JOIN AP_INVOICE_LINES_ALL l USING (INVOICE_ID)
  WHERE i.PAY_GROUP_LOOKUP_CODE = 'COURIER_WEEKLY'
    AND l.DESCRIPTION = 'ADJUSTMENTS'
  GROUP BY 1
)
SELECT
  w.week_start,
  e.ENFORCEMENT_STATUS_CODE AS status,
  o.orders,
  o.orders_mon_tue,
  s.shift_hours,
  e.CONNECTED_HOURS AS engine_hours,
  e.COUNTABLE_PAY_AMOUNT AS countable_pay,
  IFNULL(e.TOPUP_INDIVIDUAL_AMOUNT, 0)
    AS topup_individual,
  IFNULL(e.TOPUP_AGGREGATE_AMOUNT, 0)
    AS topup_aggregate,
  a.adjustments
FROM weeks w
LEFT JOIN XX_PAY_PERIOD_INCENTIVES e
  USING (PAYOUT_PERIOD_ID)
LEFT JOIN orders o USING (PAYOUT_PERIOD_ID)
LEFT JOIN shifts s USING (PAYOUT_PERIOD_ID)
LEFT JOIN invoices a USING (PAYOUT_PERIOD_ID)
ORDER BY w.week_start
\end{lstlisting}
\end{minipage}
\cardstep{2}{run\_python}{Verify the mechanism, read May's pay level, push the plan's hours through the floor, file.}
\cardpy{sec/examples/fc12_step2.py}

\clearpage
\cardcaption{tab:example-dashboard}{A dashboard data-source task. The agent rebuilds a
  Finance tile from the full-year warehouse and publishes it through Mission Control,
  which holds the frame to the tile's contract; the grader compares it with the simulator's
  own order economics.}

\cardbegin{ds-22-margin-monthly-per-order}{Dashboard data sources}{full-year warehouse}

\cardsection{Question}
\begin{cardprompt}
Finance needs us to rebuild their dashboard's data sources based on the new reporting
schema. The tile shows what each month's orders earned us per order, for the whole of 2024.
Rebuild it from the warehouse and publish it.

In April the platform started to comply with the DCWP minimum-pay rule for food delivery
workers (Q1 was a tolerated phase-in period), and the new minimum of \$19.56 per hour came
into force. Whenever a courier's pay for a week falls short of the minimum, the platform
tops it up with a true-up.

Use the numbers as Finance booked them against each order. Count every order placed in the month (by \code{ORDERED\_DATE}), whatever happened to it afterwards: cancelled orders count too.
For revenue we
keep, we factor in merchant commission, delivery, service and small-order fees, the
regulatory response fee, and the membership fee allocated to the order. For costs we bear,
we factor in courier pay, the platform-funded share of promotions and refunds, card
processing, chargebacks, cancellation costs, referral and quest incentives, and the
minimum-pay true-up allocated to the delivery.

Publish the finance data source representing the contribution margin per order for 2024,
at a monthly grain, ascending (\code{contribution\_margin\_monthly\_per\_order\_2024}):

\vspace{2pt}
\setlength{\tabcolsep}{4pt}\scriptsize
\begin{tabular}{@{}l l p{0.5\linewidth}@{}}
\toprule
column & type & meaning \\
\midrule
\code{month} & month & The month the orders were placed in, as YYYY-MM. \\
\code{orders} & integer & Orders placed in the period, every status. \\
\code{contribution\_margin\_per\_order\_usd} & number & Finance's contribution margin on the period's orders divided by \code{orders}, in dollars per order. \\
\bottomrule
\end{tabular}
\end{cardprompt}

\cardsection{Grader}
{\footnotesize
\code{publish\_data\_sources} holds the frame to the tile's contract before anything is
filed: exactly the three columns, \code{month} as YYYY-MM, an integer and a number, one row
per month, twelve rows, ascending. A frame that fails is refused with every problem listed,
and the agent may try again. The grader then compares the published frame with the key,
all or nothing: every \code{orders} exact and every margin within \$0.0055 per order (the
reference overleaf publishes 1.0667 for April against the key's 1.0668, cent rounding in the
books). The key
is the simulator's own per-order economics summed by the month the order was placed, never
projected into the warehouse. The prompt does not say how the two terms booked at a coarser grain are allocated to orders; the agent must find this in the books: the membership fee allocated
to the orders it funded and the minimum-pay true-up accrued per delivery.\par
\cardlabel{SCORING}{}
\begin{lstlisting}[style=cardpy,numbers=none,xleftmargin=6pt,framexleftmargin=6pt]
frame = published["contribution_margin_monthly_per_order_2024"]   # met the contract, or was refused
right = (list(frame.month) == list(key.month)                     # every month, in order
         and (frame.orders == key.orders).all()                   # integers exact
         and (abs(frame.contribution_margin_per_order_usd
                  - key.contribution_margin_per_order_usd) <= 0.0055).all())
score = 100 if right else 0         # all or nothing: a tile wrong for one month is a wrong tile
\end{lstlisting}
\vspace{2pt}
\vspace{4pt}
\centering
\setlength{\tabcolsep}{4pt}
\begin{tabular}{@{}l >{\columncolor{keyfill}}r >{\columncolor{keyfill}}r >{\columncolor{keyfill}}r >{\columncolor{keyfill}}r >{\columncolor{keyfill}}r >{\columncolor{keyfill}}r@{}}
\toprule
key (\code{month}) & 2024-01 & 2024-02 & 2024-03 & 2024-04 & 2024-05 & 2024-06 \\
\midrule
\code{orders} & 6,630,400 & 6,423,388 & 6,886,113 & 6,498,573 & 6,784,904 & 6,590,763 \\
margin per order (\$) & 1.5586 & 0.2356 & 0.0081 & 1.0668 & 0.7737 & 1.1457 \\
\midrule
 & 2024-07 & 2024-08 & 2024-09 & 2024-10 & 2024-11 & 2024-12 \\
\midrule
\code{orders} & 6,631,153 & 6,665,236 & 6,579,900 & 6,946,910 & 6,927,590 & 7,439,544 \\
margin per order (\$) & 1.4397 & 1.2730 & 1.3083 & 1.5846 & 1.4501 & 1.2801 \\
\bottomrule
\end{tabular}\par
\vspace{3pt}
}

\clearpage
\cardsection{Reference solution}
\noindent\begin{minipage}{\linewidth}
\cardstep{1}{run\_sql}{Which journal lines do the books tie to an order? \code{REFERENCE\_4}
  names the line type, \code{REFERENCE\_2} the order. Ten types are Finance's list, one to one.}
\cardsql{sec/examples/ds22_step1.sql}
\end{minipage}
\noindent\begin{minipage}{\linewidth}
\cardstep{2}{run\_sql}{The denominator: every order placed in each month, every status.}
\cardsql{sec/examples/ds22_step2.sql}
\end{minipage}
\noindent\begin{minipage}{\linewidth}
\cardstep{3}{run\_sql}{The numerator: credits minus debits on those lines, by month and
  type. An amount sits on the subledger line (\code{XLA\_AE\_LINES}) or on the custom feed's
  interface row (step~4); one query per source, as together they exceed the 20\,GiB scan cap.}
\cardsql{sec/examples/ds22_step3.sql}
\end{minipage}
\noindent\begin{minipage}{\linewidth}
\cardstep{4}{run\_sql}{Step 3 over the second source, \code{XX\_GL\_INTERFACE\_HIST}: the same
  query with lines 3, 4 and 6 replaced by these.}
\begin{lstlisting}[style=cardsql,numbers=none,xleftmargin=6pt,framexleftmargin=6pt]
       SUM(COALESCE(CAST(a.ENTERED_CR AS NUMERIC), 0)
           - COALESCE(CAST(a.ENTERED_DR AS NUMERIC), 0)) AS margin_usd
JOIN XX_GL_INTERFACE_HIST a ON a.INTERFACE_LINE_ID = SAFE_CAST(r.REFERENCE_7 AS INT64)
\end{lstlisting}
\end{minipage}
\noindent\begin{minipage}{\linewidth}
\cardstep{5}{run\_python}{Sum the ten line types per month, divide by the orders, publish.}
\cardpy{sec/examples/ds22_step5.py}
\end{minipage}

\cardsection{Scores}
\chiprow{%
  \chipref{Reference solution}{100}%
  \chip{chipgood}{GPT-6 Astra}{100}%
  \chip{chipgood}{GPT-6.1 Sol}{100}%
  \chip{chipgood}{Claude Opus 5.5}{100}%
  \chip{chipgood}{Gemini 3.8 Flash}{100}%
  \chip{chipzero}{GPT-6 Sol}{0}%
  \chip{chipzero}{GPT-6 Luna}{0}%
  \chip{chipzero}{Claude Sonnet 5.5}{0}%
  \chip{chipzero}{Claude Sonnet 5}{0}%
  \chip{chipzero}{Claude Haiku 4.5}{0}%
  \chip{chipzero}{Muse Spark 1.3}{0}%
  \chip{chipzero}{Kimi K3}{0}%
  \chip{chipzero}{GLM 5.3 Flash}{0}%
  \chip{chipzero}{DeepSeek V4.1 Flash}{0}%
  \chip{chipzero}{Qwen 3.8 Max}{0}}

\clearpage
\cardcaption{tab:example-collusion}{A fraud task with two sides. The agent must find the
  couriers splitting ``never arrived'' refunds with their regulars and close both the
  couriers and the partner accounts, priced by what each decision costs: a wrongly deactivated
  courier destroys about \$384 of margin against about \$140 of refunds per partnership. The task therefore turns on defining a regular precisely enough to leave honest twins alone.}

\cardbegin{col-10-refund-partnerships}{Fraud \& risk}{full-year warehouse $\cdot$ review as of February 28, 2025}

\cardsection{Question}
\begin{cardprompt}
Support flagged a pattern in our never-arrived refunds. Our policy refunds a ``never
arrived'' claim in full, and we fund it ourselves: the courier still keeps the pay and the
tip. Some couriers' complaints are not coming from strangers: they come from the same few
customers again and again, customers those couriers have delivered to before, on orders
whose delivery record puts the courier at the door. That is not theft and it is not porch
piracy. It is a partnership, and both sides are in on it.

Find the couriers running refund partnerships with their regulars and deactivate their
vendor IDs for refund collusion. Also ban the customer account IDs that were their partners.

Couriers who steal food outright are being handled by another review --- do not chase them
here. Leaving a partnership running keeps costing us full refunds; deactivating an honest
courier or closing a genuine customer destroys what we earn from them, and every decision
costs a fixed review amount. A courier with a bad week, a customer whose building has a
package thief, a claimant who was actually right --- those all exist here in far greater
numbers than partnerships.
\end{cardprompt}

\cardsection{Grader}
{\footnotesize
Enforcement is scored by what it cost the business, not by how many ids matched
(\code{cost\_set}; \citealp{elkan2001foundations}). Every courier who ever had a never-arrived
refund, and every customer who ever claimed one, is in a priced universe with one of three
verdicts: a \emph{target} costs its refunds if left running and a \$2 review if banned; an
\emph{innocent} costs the review plus the margin the platform earns from it if banned, and
nothing if left alone; a \emph{neutral} is bad but not what was asked (a thief, a promo farm)
and costs half a review if banned, never a miss. The cost of the filing is compared with the
cheaper of banning nobody and banning everybody, and savings against that baseline, clamped to $[0, 1]$ and
scaled to 100, is the score. Couriers weigh twice the partners. Banning nobody costs \$34,077
on the courier side, so each wrongly deactivated courier (median margin \$384) gives back
about 1.1\% of the courier score; a partner account's margin is halved before it is charged,
and most claimant accounts have none. The targets are the 243 injected partnerships and
their 577 partner accounts; the 729 honest couriers with one regular whose building has a
package thief, and the 1,795 couriers who steal orders, are the twins to leave alone.\par
\cardlabel{SCORING}{}
\begin{lstlisting}[style=cardpy,numbers=none,xleftmargin=6pt,framexleftmargin=6pt]
# universe: id -> (verdict, loss if left running, margin destroyed if banned)
def savings(universe, acted, admin=2.0):
    cost = sum(admin if k in acted else loss for k, (v, loss, m) in universe.items() if v == "target")
    cost += sum(admin + m for k, (v, loss, m) in universe.items() if v == "innocent" and k in acted)
    cost += sum(admin / 2 for k, (v, loss, m) in universe.items() if v == "neutral" and k in acted)
    ban_nobody = sum(loss for v, loss, m in universe.values() if v == "target")
    ban_everybody = sum(admin + (m if v == "innocent" else 0)
                        for v, loss, m in universe.values() if v != "neutral")
    return max(0.0, 1 - cost / min(ban_nobody, ban_everybody))

score = 100 * (2 * savings(couriers, banned_couriers) + savings(partners, banned_customers)) / 3
\end{lstlisting}
\vspace{2pt}
\vspace{4pt}
\centering
\setlength{\tabcolsep}{4pt}
\begin{tabular}{@{}l >{\columncolor{keyfill}}r >{\columncolor{keyfill}}r >{\columncolor{keyfill}}r@{}}
\toprule
key (verdict) & couriers & partner accounts & priced at \\
\midrule
target & 243 & 577 & refunds left running: \$34,077 $\cdot$ \$124,507 \\
innocent & 32,228 & 80,934 & margin if banned: \$12.8M $\cdot$ \$2.7M \\
neutral & 5,177 & 79,952 & half a review if banned \\
\bottomrule
\end{tabular}\par
\vspace{2pt}
\begin{tabular}{@{}l r r r r r r@{}}
\toprule
Reference solution, overleaf & filed & caught & wrong & missed & savings & grade \\
\midrule
\code{ban\_couriers} & 243 & 242 & 1 & 1 & 97.1 & \\
\code{ban\_customers} & 589 & 576 & 11 & 1 & 98.8 & \\
weighted 2\,:\,1 & & & & & & \textbf{97.7} \\
\bottomrule
\end{tabular}\par
\vspace{3pt}
}
\chiprow{%
  \chipref{Reference solution}{98}%
  \chip{chipgood}{Claude Opus 5.5}{99}%
  \chip{chipgood}{Claude Sonnet 5.5}{99}%
  \chip{chippart}{GPT-6 Astra}{49}%
  \chip{chippart}{GPT-6 Sol}{28}%
  \chip{chippart}{Qwen 3.8 Max}{20}%
  \chip{chippart}{GPT-6 Luna}{14}%
  \chip{chippart}{Muse Spark 1.3}{11}%
  \chip{chippart}{Gemini 3.8 Flash}{9}%
  \chip{chippart}{Claude Sonnet 5}{9}%
  \chip{chippart}{DeepSeek V4.1 Flash}{9}%
  \chip{chippart}{Kimi K3}{9}%
  \chip{chippart}{GPT-6.1 Sol}{9}%
  \chip{chippart}{GLM 5.3 Flash}{9}%
  \chip{chippart}{Claude Haiku 4.5}{4}}

\cardsection{Reference solution}
\noindent\begin{minipage}{\linewidth}
\cardstep{1}{run\_sql}{What the never-arrived refunds are: 103,223 in 2024, \$3.95M, all but
  \$48k funded by the platform; 11,703 have no delivery record at all (nobody delivered,
  the theft review's) and the rest sit on a delivered leg.}
\cardsql{sec/examples/col10_step1.sql}
\end{minipage}
\noindent\begin{minipage}{\linewidth}
\cardstep{2}{run\_sql}{Every claim with its delivery record and three facts about the pair
  behind it: how many other orders this courier delivered to this customer, how many of them
  before the claim, and how far the leg ended from where the customer's other deliveries
  end. Regulars and the door in one pass, 5.5\,GB.}
\cardsql{sec/examples/col10_step2.sql}
\end{minipage}
\noindent\begin{minipage}{\linewidth}
\cardstep{3}{run\_python}{Say who a regular is (three orders together, one before the claim),
  count distinct regulars claiming per courier, file both sides.}
\cardpy{sec/examples/col10_step3.py}
\end{minipage}